\documentclass{article}

\usepackage{arxiv}

\usepackage[utf8]{inputenc} 
\usepackage[T1]{fontenc}    
\usepackage{hyperref}       
\usepackage{url}            
\usepackage{booktabs}       
\usepackage{amsfonts}       
\usepackage{nicefrac}       
\usepackage{microtype}      
\usepackage{lipsum}
\usepackage{xr}

\usepackage{longtable}
\usepackage[utf8]{inputenc} 
\usepackage[T1]{fontenc}    
\usepackage{url}            
\usepackage{booktabs}       
\usepackage{amsfonts}       
\usepackage{nicefrac}       
\usepackage{microtype}      
\usepackage{lipsum}
\usepackage{fancyhdr}       
\usepackage{graphicx}       
\graphicspath{{media/}}     
\usepackage{tabularx}
\usepackage{listings} 
\usepackage{fancyvrb} 

\usepackage[numbers]{natbib}
\usepackage{amsmath,amssymb}
\usepackage{amsthm}
\usepackage{mathtools}
\usepackage{xspace}
\usepackage[noend]{algorithmic}
\usepackage[ruled,vlined]{algorithm2e}
\usepackage{enumitem}

\usepackage{pdfpages}

\usepackage{chngcntr}
\usepackage{mdframed} 
\usepackage{caption}

\usepackage{color}
\usepackage{xcolor}

\usepackage{subfigure}
\usepackage[most]{tcolorbox}
\usepackage[frozencache=true, finalizecache=false, cachedir=./minted-cache]{minted}

\usepackage{float}
\usepackage{alltt}
\usepackage{soul}
\usepackage{fancyvrb}
\usepackage{multirow}

\usepackage{csquotes}

\newcounter{textbox}
\usepackage{mdframed} 

\usepackage{listings} 

\tcbset{
  custombox/.style={
    width=474.18663pt,
    top=3pt,
    fonttitle=\bfseries\small\sffamily, 
    colback=white,
    colframe=black,
    colbacktitle=black,
    boxrule=0.5pt, 
    colback=white, 
    enhanced,
    rounded corners, 
    center,
    boxed title style={
            sharp corners,
            size=small,
            colframe=blue!75!black, 
        },
    attach boxed title to top left={yshift=-0.1in,xshift=0.15in},
    boxed title style={boxrule=0pt,colframe=white,},
  }
}

\newtcolorbox{LLMbox}[2][]{custombox,title=#2,#1}

\tcbset{
  customboxmultipage/.style={
    width=474.18663pt,
    top=3pt,
    fonttitle=\bfseries\small\sffamily, 
    colback=white,
    breakable,
    colframe=black,
    colbacktitle=black,
    boxrule=0.5pt, 
    colback=white, 
    enhanced,
    rounded corners, 
    center,
    boxed title style={
            sharp corners,
            size=small,
            colframe=blue!75!black, 
        },
    attach boxed title to top left={yshift=-0.1in,xshift=0.15in},
    boxed title style={boxrule=0pt,colframe=white,},
  }
}
\newtcolorbox{LLMboxmultipage}[2][]{customboxmultipage,title=#2,#1}

\newtcbox{\mybox}[1][green]{on line,
arc=0pt,outer arc=0pt,colback=#1!10!white,colframe=#1!50!black,
boxsep=0pt,left=0pt,right=0pt,top=0pt,bottom=0pt,
boxrule=0pt,bottomrule=0pt,toprule=0pt}

\definecolor{aigold}{RGB}{244,210, 1} 
\definecolor{aigreen}{RGB}{210,244,211} 

\sethlcolor{aigreen}

\definecolor{aired}{RGB}{255,180,181}

\definecolor{lightred}{rgb}{1,0.9,0.9} 

\title{LifeGPT: Topology-Agnostic Generative Pretrained Transformer Model for Cellular Automata}

\author{
 Jaime A. Berkovich \\
 Laboratory for Atomistic and Molecular Mechanics (LAMM)\\
  Department of Materials Science and Engineering\\
  Massachusetts Institute of Technology\\
  77 Massachusetts Ave.\\
  Cambridge, MA 02139, USA \\
   \And
 Markus J. Buehler \\
 Laboratory for Atomistic and Molecular Mechanics (LAMM)\\
 Department of Civil and Environmental Engineering\\
 Department of Mechanical Engineering\\
 Center for Computational Science and Engineering\\
 Schwarzmann College of Computing,\\
  Massachusetts Institute of Technology\\
  77 Massachusetts Ave.\\
  Cambridge, MA 02139, USA \\
  \texttt{mbuehler@MIT.EDU} \\
}

\begin{document}
\maketitle
\begin{abstract}
Conway's Game of Life (Life), a well known algorithm within the broader class of cellular automata (CA), exhibits complex emergent dynamics, with extreme sensitivity to initial conditions. Modeling and predicting such intricate behavior without explicit knowledge of the system's underlying topology presents a significant challenge, motivating the development of algorithms that can generalize across various grid configurations and boundary conditions. We develop a decoder-only generative pretrained transformer (GPT) model to solve this problem, showing that our model can simulate Life on a toroidal grid with no prior knowledge on the size of the grid, or its periodic boundary conditions (LifeGPT). LifeGPT is topology-agnostic with respect to its training data and our results show that a GPT model is capable of capturing the deterministic rules of a Turing-complete system with near-perfect accuracy, given sufficiently diverse training data. We also introduce the idea of an `autoregressive autoregressor' to recursively implement Life using LifeGPT. Our results pave the path towards true universal computation within a large language model framework, synthesizing of mathematical analysis with natural language processing, and probing AI systems for situational awareness about the evolution of such algorithms without ever having to compute them. Similar GPTs could potentially solve inverse problems in multicellular self-assembly by extracting CA-compatible rulesets from real-world biological systems to create new predictive models, which would have significant consequences for the fields of bioinspired materials, tissue engineering, and architected materials design.
\end{abstract}


\section{Main}\label{sec:main}
Cellular automata (CA) have long been a subject of profound interest within the fields of computer science and mathematics, owing to their intricate and emergent behaviors. CA algorithms are uniquely characterized by their combination of computational simplicity—evolving solely by local state-transition rules—and broad dynamical behavior, encompassing static, periodic, chaotic, and complex patterns, depending on the ruleset and initial condition (IC) being used. These properties render CA algorithms particularly valuable for simulating a wide array of natural phenomena, such as the propagation of forest fires~\cite{hernandez_encinas_simulation_2007}, traffic flow dynamics~\cite{zhao_cellular_2020}, chemical reactions~\cite{noauthor_scientific_1988}, and recrystallization~\cite{hesselbarth_simulation_1991}. The inherent behavioral unpredictability of CA (following human inspection of their rulesets)~\cite{wolfram_new_2002} has hindered advancements in these subfields, confining CA to the realm of phenomenological modeling and subsequently preventing their evolution into mature, predictive tools for systems or phenomena where do not yet know a closed-form ruleset.

\begin{figure}[!h]
\centering
\includegraphics[width=0.5\columnwidth]{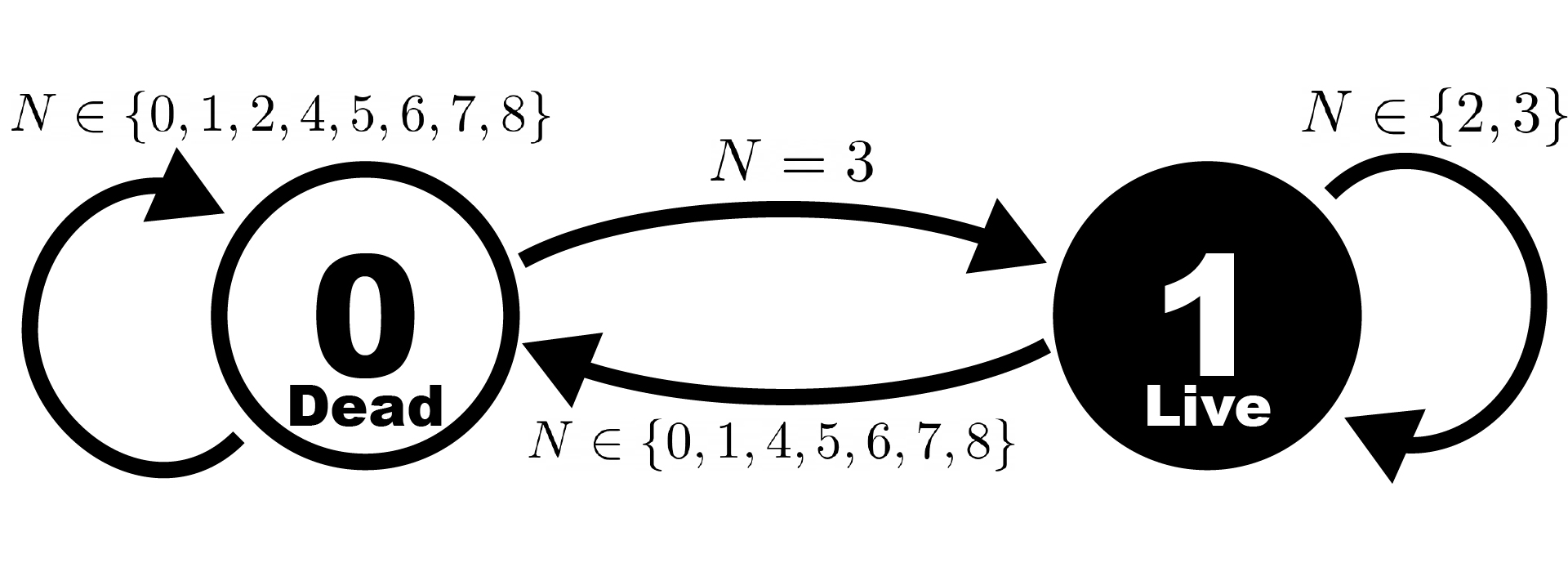}
\caption{A state transition diagram for a single cell in Life. \(N\) represents the number of neighboring cells in state 1 (live cells). The two large circles represent the possible states of a cell. Arrows represent deterministic state transitions.}\label{fig:Life_state_diagram}
\end{figure}

One notable CA algorithm is Conway's Game of Life (Life), as introduced by John H. Conway in 1970~\cite{gardner_mathematical_1970}. Life (also referred to as simply `the Game of Life') has since intrigued researchers from a range of disciplines, including mathematics, computer science and materials science. This is partly due to Life's self-organizing yet often unpredictable dynamics~\cite{bak_self-organized_1989} that emerge from a simple 2-state-8-neighbor transition scheme (Fig.~\ref{fig:Life_state_diagram}). Despite Life's fame, complex dynamics are present in numerous CA, and have led many researchers to support the conjecture that most CA are `computationally irreducible.' This designation suggests that their evolution cannot be perfectly predicted by any process more computationally efficient than running the CA algorithms themselves~\cite{wolfram_new_2002}. Practically, this conjecture implies that the evolution of most CA algorithms from an arbitrary IC to an arbitrary time-step in the future cannot be described analytically. As Wolfram (2024)~\cite{wolfram_can_2024} writes, ``... we can't expect to systematically `jump ahead'' and predict [...] the system...'' Thereby, any model that tries to simulate or predict the dynamics of a computationally irreducible CA is either implementing an algorithm capable of approximating CA behavior, or implementing the exact CA algorithm as a recursive operation for the number of timesteps desired. The former solution requires that the model diverge from the CA algorithm's ground truth after some number of timesteps and/or for some specific initial conditions, and the latter solution necessitates that the model be fundamentally limited by its size to a maximum number of timesteps. Another question is whether, and how, we can begin to build artificially intelligent systems that can program new versions of CA to meet certain target behaviors, and whether such a system could develop a heightened situational `awareness' to predict the evolution of an algorithm multiple timesteps into the future without explicitly computing it recursively. While this may seem impossible given the computational irreducibility conjecture, past research has suggested that Life exhibits self-organized-criticality (SOC)~\cite{bak_self-organized_1989} and persistent scaling behavior~\cite{fehsenfeld_persistence_1998}. Additionally, some CA have been shown to generate fractal-like patterns~\cite{martin_inherent_1994}. These phenomena suggest that Life (and potentially, many other CA algorithms~\cite{wolfram_statistical_1983}) are at least statistically predictable -- while perfectly accurate predictions of any CA system might be unachievable without running the exact CA algorithm, being able make global or `coarse-grained' claims about CA systems is far more reasonable. In addition, recent work~\cite{zhang_intelligence_2024} demonstrates that exposing LLMs to complex elementary CA systems (class 4~\cite{wolfram_new_2002}), as opposed to static (class 1), periodic (class 2), or chaotic (class 3) CA systems, leads to improved downstream performance on cognitive tasks, suggesting that a fundamental understanding of complexity is intrinsic to the common notion of `intelligence'. The nexus question here is: Are there `deep trends' (which may also be conceptualized as `pockets of computational reducibility'~\cite{wolfram_can_2024}) that exist across, or transcend, scales in CA that might be elucidated by deep learning models? These and related questions led us to explore the use of AI algorithms in modeling CA.

With increases in computing power and the growing popularity of neural networks as tools of analysis for complex systems (for which there are few practically useful analytical models), the computational irreducibly problem has lead many down the road of using neural networks to try to `predict' the evolution of CA systems (often, Life is chosen as the example system), with varying degrees of success. The general assumption across this field of research is that neural networks can either learn the rules of a given CA algorithm exactly, or can learn to abstract the system's behavior well enough to find a more-or-less accurate prediction of some far-future state based on an IC, though some work ventures beyond this paradigm. 

Previous work focuses largely on either feed-forward convolutional neural networks (CNNs)~\cite{springer_its_2021} or convolutional encoder-decoder models~\cite{aach_generalization_2021}. Such approaches already encode knowledge of the significance of the spatial relationships between each cell in the CA training data, namely that CA utilize state-transition rules based on the neighboring states of a given cell (also known as the Moore neighborhood~\cite{weisstein_moore_nodate,gray_mathematician_2003}). This likely introduces a form of inductive bias. CNNs embed the solution in the model directly because they already understand that CA utilize a 2D grid as a logical topology and hence even before the their weights are initialized, CNNs are more than half way to the solution. To the best of our knowledge, no research has investigated the possibility of utilizing a data topology-agnostic model for predicting the evolution of CA.

Within the past years, generative pretrained transformer models (GPTs), have gained widespread popularity within the areas of natural language processing~\cite{gillioz_overview_2020}, weather prediction~\cite{nguyen_scaling_2023, alerskans_transformer_2022, saleem_stc-vit_2024}, speech processing/modelling~\cite{latif_transformers_2023, mamatov_speech_2021, li_neural_2019}, machine vision~\cite{berroukham_vision_2023}, strategic gaming~\cite{xu_language_2024, hu_survey_2024, liu_large_2024}, physical field modelling~\cite{buehler_fieldperceiver_2022}, protein folding~\cite{moussad_transformative_2023} and finance~\cite{brugiere_transformer_2024, lezmi_time_2023}, to name a few. This remarkable capability for GPTs to generate accurate predictions, solve inverse problems, make useful decisions, and/or new and relevant content depending on the task at hand has been attributed to the underlying attention mechanism's remarkable ability to parse meaning from data through repeatedly updating token embeddings in a high-dimensional space~\cite{vaswani_attention_2023}. Consequentially, attention allows GPTs to build sophisticated ontological understandings, which must be learned using a large amount of training data~\cite{saeedizade_navigating_2024, buehler_generative_2024, buehler_accelerating_2024}, or fine-tuned (in the case of pretrained models) on a smaller set of data~\cite{luu_bioinspiredllm_2023}. Furthermore, these understandings can be visualized in numerous ways using graphs, proving a unique level of functional transparency~\cite{buehler_generative_2024,buehler_accelerating_2024}, in contrast to previous machine learning strategies, that are more opaque, leading to the common `black-box' comparison~\cite{dobson_reading_2023}.

In this paper we present an analysis of the transformer architecture's capability to predict the outcomes of Life, played on a 32 $\times$ 32 toroidal grid, with remarkable accuracy. We accomplish this using a decoder-only transformer model, equipped with causally masked multi-headed self-attention, with forgetful causal masking (FCM) implemented during training, trained on data representing pairs of 2D grids encoding ICs and next-game-states (NGSs); we call our model LifeGPT.

\paragraph{Transformer Models and Artificial Life}
ALife is a highly interdisciplinary field that is focused on understanding the essential attributes of life and life-like systems through computation, hardware, and biochemical means. `Soft' ALife, which is the most relevant to our work, is a subset of ALife that is entirely simulated computationally, instead of being physically realized~\cite{bedau_artificial_2007}. Many CA systems (notably, Life), have been constructed for the purpose of better understanding self-assembly, self-propagating patterns, and, more generally, complexity and chaos, all phenomenological attributes intrinsic to life as it is currently understood~\cite{zhao_progressive_2013,capra_complexity_2005,skinner_application_1996,kurakin_self-organizing_2011}.

Some scholars have already begun to conceptualize GPTs -- more specifically, large language models (LLMs) -- as having a `reciprocal relationship' with ALife research. Nisioti \textit{et al.} (2024)~\cite{nisioti_text_2024} identify two key relationships between the two areas of research. The authors argue that LLMs may become tools for ALife research by allowing for the generation of open-ended environments, or as operators of evolutionary computation. Inversely, they also argued, ALife principles of such as collective intelligence, evolution, and organization may also be exhibited by LLM agents. The authors emphasize the potential for LLMs for control or understand ALife systems, which is the primary focus of our paper. While the scope of our paper is limited to a replication of Life's game-state transition rules using a transformer model, our work lays the groundwork for a paradigm in which GPTs are able to fine-tune CA systems' emergent qualities through alterations of ICs and rulesets, enabling a form of ALife oversight and regulation, which the authors argue is a primary ethical consideration.

\paragraph{Cellular Automata as Discrete Neural Network Models}
In recent months, yet another reciprocal relationship between neural networks and ALife has been elucidated. Recent work by Wolfram~\cite{wolfram_whats_2024} shows that discrete systems known as `spacetime-inhomogeneous cellular automata' (also referred to as `rule arrays') can be tuned through a discrete, Boolean-calculus-based backpropogation to create minimal models for perceptron-style neural-networks. In addition, it is argued that ordinary cellular automata can, similarly, discretely model recurrent neural networks. Through these examples, an argument is developed asserting that modern-day machine learning techniques are effective at learning simply because they can `mine' programs from an extremely large space of complex behavior, arising from computationally irreducible systems governed by relatively few/simple rules. Thus, the author  argues, it is not likely that AI will intrinsically favor finding `explainable' or `understandable' models for fitting training data, rather, AI will find something that just ``happen[s] to work''. Furthermore, it is stated that in the context of AI development, this harnessing of computational irreducibly creates a fundamental trade-off between model interpretability/predictability and model capability. Models that are structured more obviously for human understanding will be more computationally reducible, and will therefore be forced to sample from a computational space with reduced complexity during weight-tuning. Moreover, it is suggested that models will intrinsically struggle to fit data generated by computationally irreducible programs, as finding programs that ``do more or less the right thing'' will not be sufficient. The work suggests that a deeper understanding of cellular automata will go hand-in-hand with the development of any kind of broadly applicable AI-science, as both necessitate the discovery and analysis of `pockets' of computationally reducible behavior inside of complex systems.

\paragraph{Convolutional Neural Networks}
Earlier work synthesizing the domains of CA and ML has largely focused on convolutional models paired with 2D CA. This is because 2D CA `time-slices' are easily represented as images, which feed-forward CNNs are especially equipped to process. Springer and Kenyon (2021)~\cite{springer_its_2021} conducted a thorough empirical analysis on the ability of feed-forward CNNs to capture the allegedly computationally irreducible dynamics of Life. Since Life abides by rules pertaining to the states of 2D nearest-neighbors (the $r=1$ Moore neighborhood), which are effectively equivalent to a convolutional operations, the authors were able to calculate a hypothetical minimal feed-forward CNN (using ReLU activation) capable of capturing the rules of the Life and applying them over $n$ time-steps. They were thereby able to define a metric for network over-completeness, $m$, to empirically test the relationship between CNN size and Life learning effectiveness for various $n$ and $m$. One finding was that for $n\geq3$, increasing $m$ up to 25 was insufficient for reaching a training convergence fraction of over 50\%. Furthermore, the authors found that for networks which did converge, the number of epochs necessary increased substantially with $n$. The authors also note that converged, minimal networks were highly sensitive to sign-flipping of initial weights, suggesting that for smaller networks, more luck in initializing suitable weights is needed to converge. The authors further reported that gradient descend was highly sensitive to the distribution parameters of the dataset, suggesting that some game examples were more useful that others for teaching the rules of Life to the model. 

\paragraph{Sigma-Pi Networks}
Wulff and Hertz (1992) demonstrated the use of simple neural networks with short range connections (\(\Sigma-\Pi\) networks) to learn the dynamics of 2D CA~\cite{wulff_learning_1992}. They concluded that even with a network sharing the same topology as the CA being studied, the dynamics of Life could not be learned effectively without weight-sharing between neurons. Their early approach was similar to many modern-day graph neural network architectures. Still, their approach encoded part of the solution to the problem in the architecture of the model, and in doing so introduced an inductive bias likely aided the learning process. By enabling weight-sharing and employing short-range connections, the architecture of the network was itself a reflection of the grid-topology and nearest-neighbor rules engendering Life.

\paragraph{Neural CA}
Recent work has shown that CA systems can be trained to dynamically sustain desired patterns (such as small 2D images), by allowing a single CNN to simultaneously control the state-transition rules of all cells in the system, which is likewise made possible by allowing cells to take on continuous states (as opposed to discrete states). By training the neural network (through repeated runs of the CA growth process, followed by subsequent loss calculation by comparison with the desired image and backpropagation), the model's parameters are tuned so that a single cell can, over time, multiply into a cellular collective that `grows' into the desired image~\cite{mordvintsev_growing_2020}.

\paragraph{Paper Outline}
The plan of the paper is as follows. First, we introduce the LifeGPT architecture and discuss the effectiveness of the model under a variety of training and inference strategies, draw conclusions regarding the manner in which the LifeGPT learns, and discuss zero/few-shot learning capabilities of LifeGPT.  Next, we introduce the concept of the `autoregressive autoregressor' (ARAR), which allows LifeGPT to recursively simulate Life's dynamics over multiple time steps. Finally, we discuss the broader implications of LifeGPT for ALife, universal computation within an LLM framework, and solving inverse problems for the design of biomimetic and bioinspired materials. Looking ahead, we provide an argument for why enhancing LifeGPT with reinforcement learning (RL) techniques could improve inference accuracy, extend its abilities to a wide space of CA rulesets, and further enhance its utility in simulating and understanding complex, life-like systems.

\section{Results and Discussion}\label{sec:results}

\subsection{Model Convergence During Training}
Details on the implementation of the model, training and datasets used are provided in Materials and Methods (section \ref{sec:methods}). LifeGPT is architected as a decoder-only transformer model with causally-masked self attention -- in addition to FCM during training -- which utilized 12 transformer layers and 8 attention heads to capture complex patterns in Life. The model's maximum sequence length were configured to accommodate a data corresponding to ICs and NGSs. Rotary positional embedding (RPE) was used to maintain spatial awareness, while the Adam optimizer and cross-entropy loss (CEL) function guided the training process (see section \ref{sec:hyper} for a full list of hyperparameters). The best-performing version of LifeGPT was trained on training data with broad-entropy ICs, though using high-entropy ICs was also tested (see sections \ref{sec:train_set_gen} and \ref{sec:order_effects}). 

\begin{figure}[H]
    \centering
    \includegraphics[width=\columnwidth]{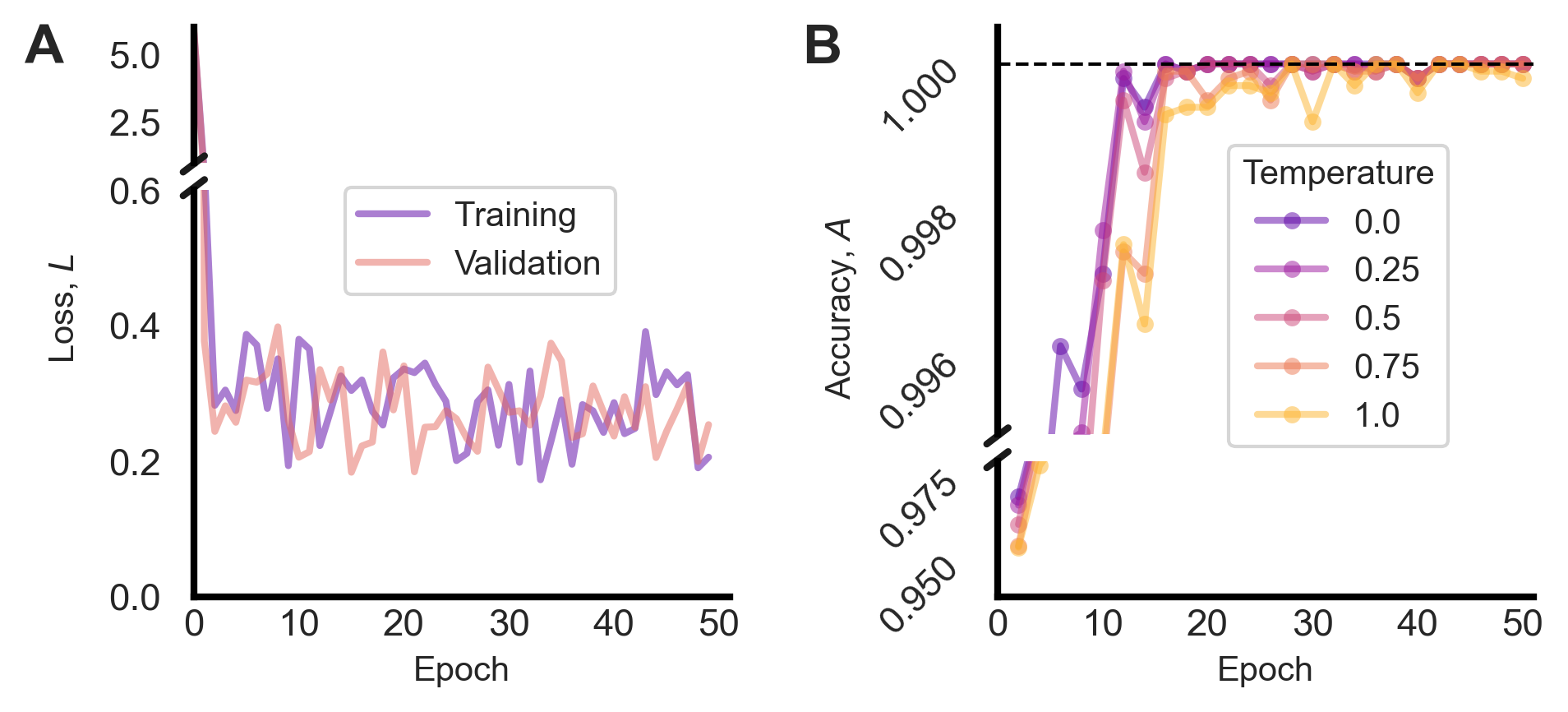}
    \caption{Multiple model performance benchmarks vs. epoch. (A) Training and validation losses. (B) Accuracy across varying temperatures.}
    \label{fig:Model_Performance}
\end{figure}

Our final model displayed rapid convergence to a range of cross-entropy loss (CEL) values from approximately 0.4 to 0.2 (Fig. \ref{fig:Model_Performance}A). We suspect the reason the loss stays far greater than 0 is that there is no causal relationship between earlier tokens and later tokens in ICs within the training data, since ICs were generated stochastically (see section \ref{sec:train_set_gen}). Since FCM was the only masking strategy used (see section \ref{sec:FCM} for details on implementation and Supplementary section \ref{sec:maskingeffects} for brief discussion of this strategy), the training and validation losses reflected the inability of our  model to predict ICs, even if the model did quickly learn to  predict NGSs from ICs (Fig. \ref{fig:Model_Performance}B).

\subsection{Accuracy Benchmarking and Sampling Temperature Effects}\label{sec:benchmarking}
\begin{figure}[H]
    \centering
    \includegraphics[width=\columnwidth]{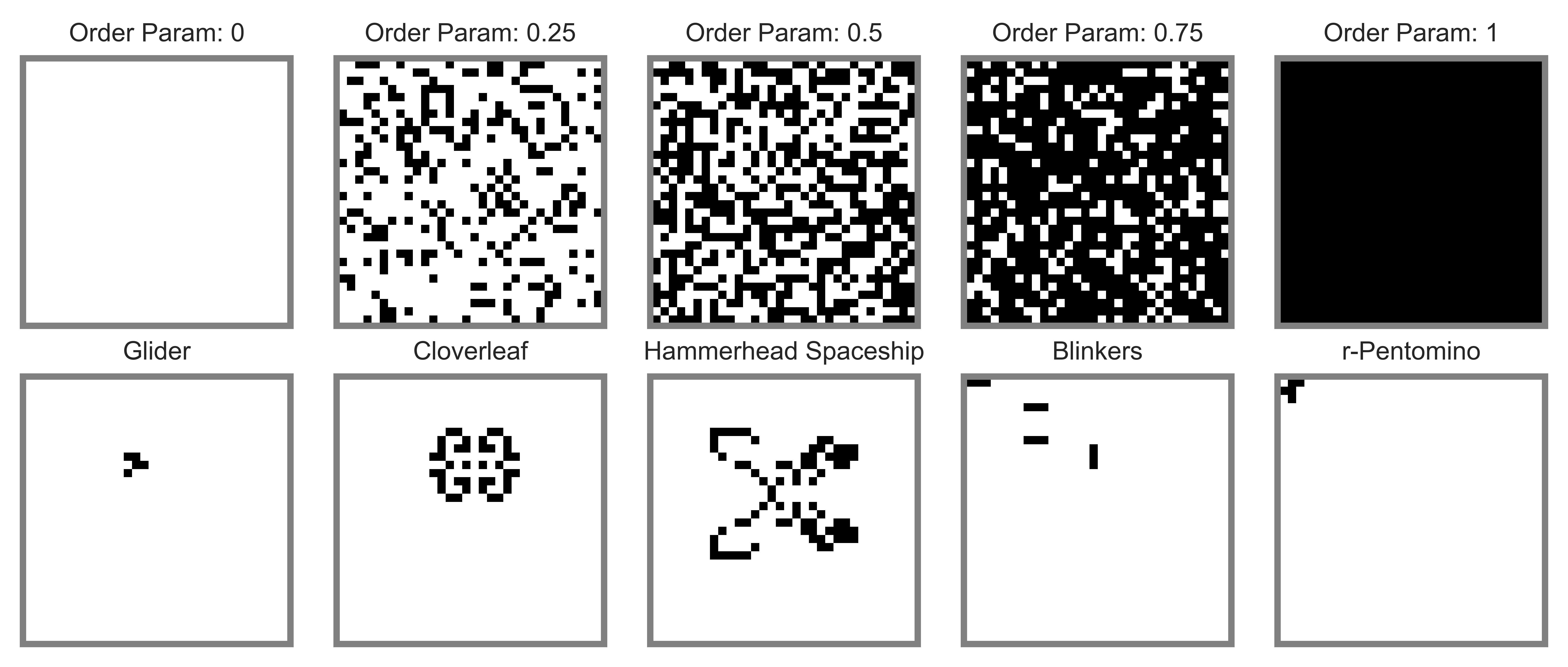}
    \caption{ICs in the test set. Top row: ICs with increasing \(\eta\) from left to right. Bottom row: ICs corresponding to known periodic and complex patterns in Life.}
    \label{fig:testing_set}
\end{figure}

CEL minimization was used during model training, but performance was assessed with periodic benchmarking. Specifically, the model was periodically tasked with autoregressively generating the tokens following 10 prompts, each containing an IC from the testing set (see section \ref{sec:testing_set_gen} and Fig. \ref{fig:testing_set}). The number of tokens correctly predicted by the model, across 10 ICs in the testing set was used to determine an accuracy score (see section \ref{sec:model_train} and equation \ref{eq:acc}).

We observed temperature does have a substantial impact on accuracy (Fig. \ref{fig:Model_Performance}B). Even though the model was trained on deterministic data, it was prone to some minor errors, especially as sampling temperature is increased. Nevertheless, even with errors, models across all temperatures tested (0.0, 0.25, 0.50, 0.75, 1.0) still achieved at least 99.9\% accuracy after about 20 epochs. Moreover, the variability in accuracy across temperatures was observed to decrease as the number of epochs of training were increased, suggesting that training for more epochs could have further mitigated errors. For this task, a temperature of 0.0 showed the best empirically-measured accuracy scores across nearly all epochs. After epoch 16, the model rarely failed to achieve 100\% accuracy, although there were couple exceptions where the model missed one or two tokens. Indeed, even for the same model trained for 50 epochs, there was one sample in the testing set, `r-pentomino' (see \href{https://github.com/lamm-mit/LifeGPT/blob/main/Testing_Set_ARAR_Animations_and_Figures/Animations/Timesteps_10/Epoch%2050/Temperature%200/evolution_temp_0_sample_10.gif}{r-Pentomino Animation at Temperature 0 -- 10 Timesteps}), for which LifeGPT's recursive NGS predictions failed to agree with the ground truth, due to a single incorrectly predicted cell early on.

\subsection{Training Data Ordering Effects}\label{sec:order_effects}
\begin{figure}[H]
    \centering
    \includegraphics[width=\columnwidth]{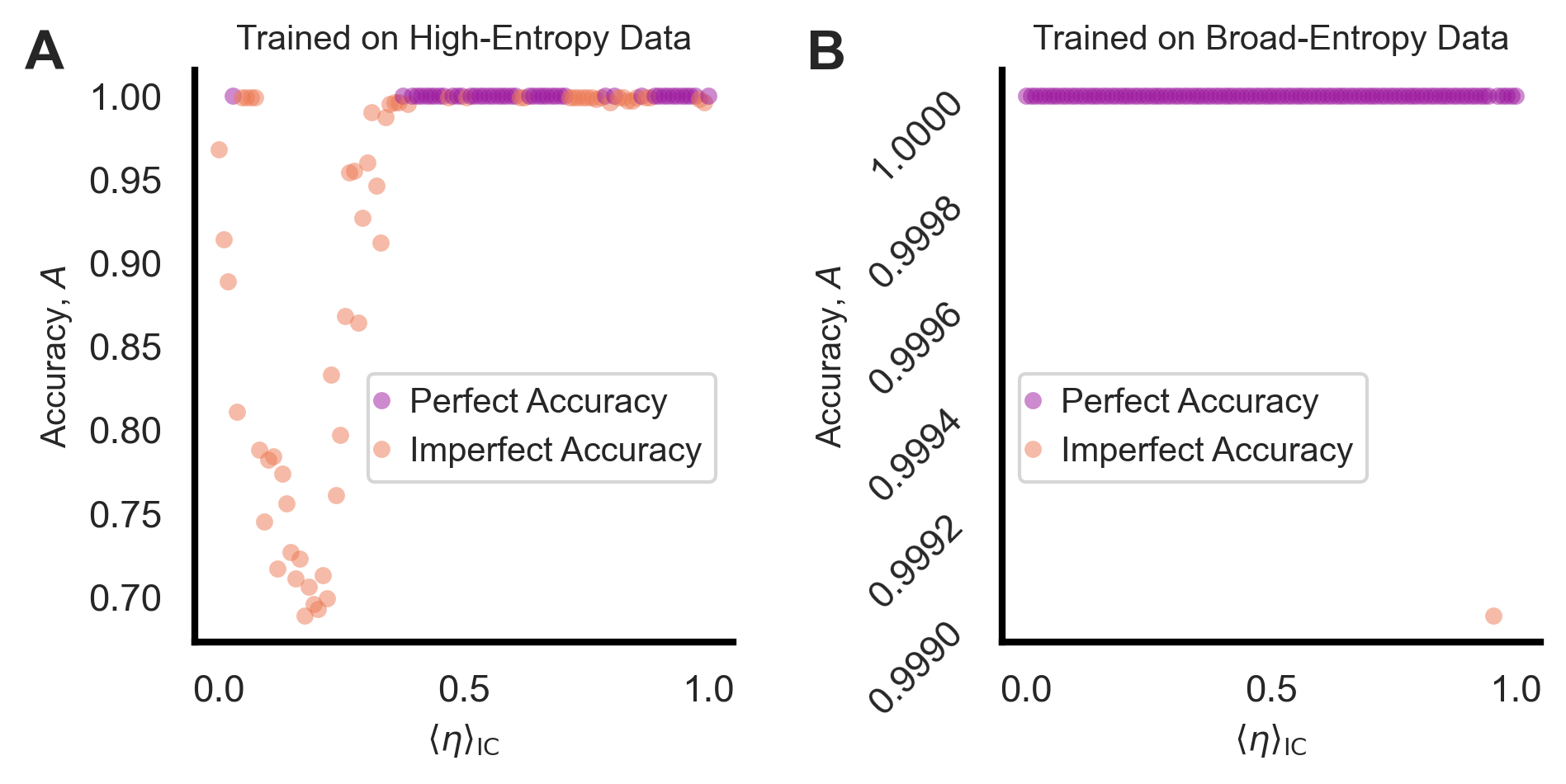}
    \caption{Model output accuracy, \(A\), versus expected order parameters of inputted ICs, \(\langle\eta\rangle_{\mathrm{IC}}\), for identical LifeGPT models trained for 50 epochs on two different training sets. Both models were set to a temperature of 1 for this experiment. (A) LifeGPT trained on a dataset with high-entropy ICs (\(\langle\eta\rangle\)=0.5). Note that this model was also trained without FCM. (B) LifeGPT trained on a dataset with broad-entropy ICs (\(\langle\eta\rangle\in[0,1]\)).}
    \label{fig:acc_ranging_order_param}
\end{figure}

Similar to earlier work using CNNs~\cite{springer_its_2021}, we find that the ordering of the samples found in training data (see section \ref{sec:train_set_gen}) has a substantial effect on model accuracy (Fig. \ref{fig:acc_ranging_order_param}). However, while earlier work used convergence (defined as achieving a critically low CEL) as a performance metric, we used accuracy (see sections \ref{sec:benchmarking} and \ref{sec:model_train}, and equation \ref{eq:acc}).

For the version of LifeGPT trained on high-entropy IC data, accuracy was reasonably high (\(A>99.5\%\)), but imperfect for samples where \(\langle\eta\rangle_{\mathrm{IC}}\in\sim[0.4,1]\). In contrast, for samples where \(\langle\eta\rangle_{\mathrm{IC}}\in\sim[0.1,0.3]\), the model performed poorly, even dipping into the \(A<80\%\) range. Intriguingly, the model regained some of its Accuracy for \(\langle\eta\rangle_{\mathrm{IC}}\in\sim[0,0.1)\) (Fig. \ref{fig:acc_ranging_order_param}A). These results suggest that the model learned some general trends related to large regions of 0-state (dead) cells resulting in more 0-state cells in the NGS, even if it did not learn how to handle ICs with mostly 0s but still a considerable fraction of 1s.

For the version of LifeGPT trained on broad-entropy samples, perfect accuracy was observed for nearly all tested ICs, regardless of \(\langle\eta\rangle_{\mathrm{IC}}\) (Fig. \ref{fig:acc_ranging_order_param}B). There was a single outlier for \(\langle\eta\rangle_{\mathrm{IC}}\approx0.954\), which we attribute to model temperature effects, since this experiment was conducted with temperature  \(=1\). Based on these results, we argue that LifeGPT effectively learned the rules for Life -- this is further supported by our experiments on our custom testing set (see section \ref{sec:benchmarking} and Fig. \ref{fig:Model_Performance}B).

\begin{figure}[H]
    \centering
    \includegraphics[width=\columnwidth]{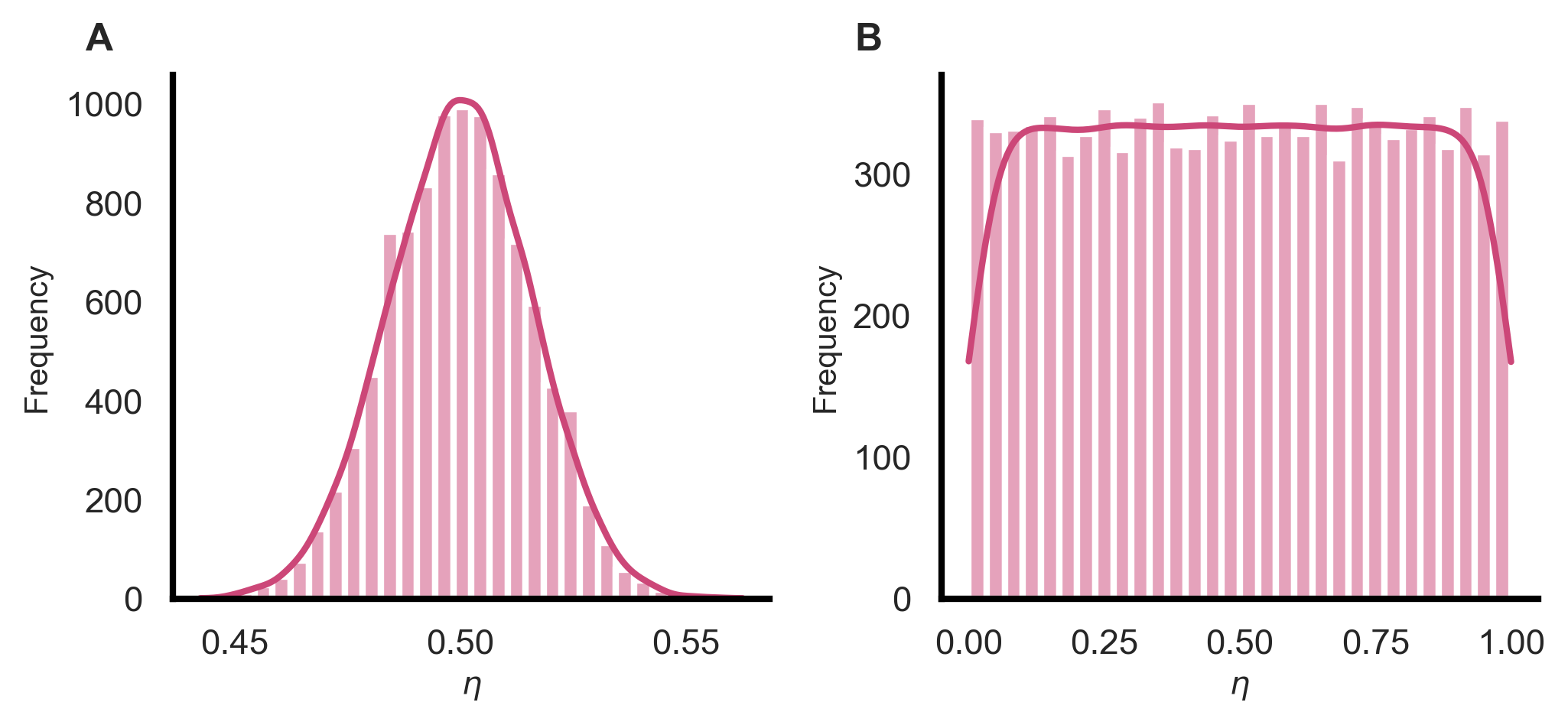}
    \caption{Frequency of an IC in the training set vs. measured IC order parameter (\(\eta\)). (A) High-entropy data generated with \(\langle\eta\rangle=0.5\). (B) Broad-entropy data generated with \(\langle\eta\rangle\) ranging from 0 to 1.}
    \label{fig:order_param_dist}
\end{figure}

While the difference in the performance of both models is stark, these findings allow us to draw two significant conclusions. The first is that the fundamental state-transition asymmetry in Life (state-0 and state-1 cells do not follow reciprocal rulesets) impacted the learning process. It appeared that it was `easier' for the model to learn how to predict NGS when many 1-state cells were in the IC, suggesting that the `overcrowding' and `stay living' rules (see the lower-central and rightmost arrows in Fig. \ref{fig:Life_state_diagram}) was more learnable than the `birth' rule (see the upper-central arrow in Fig. \ref{fig:Life_state_diagram}). The return to reasonably high accuracy for \(\langle\eta\rangle_{\mathrm{IC}}\in\sim[0,0.1)\) suggests that the `stay dead' rule (see the leftmost arrow in Fig. \ref{fig:Life_state_diagram}) was also reasonably learnable. One possible reason for the model's inability to extrapolate the birth rule could be that this rule has the most specific conditions for being enacted, requiring a 0-state cell to have exactly 3 1-state neighbors -- for a high-entropy training set, the model is simply going to encounter instances of the other 3 rules more frequently. As shown in Fig. \ref{fig:order_param_dist}A, there was still some variation in the measured order parameters of samples in the training set (the distribution was binomial because the data generation process was stochastic in the same way as flipping a fair coin -- see section \ref{sec:train_set_gen}). Still, this variation was clearly not enough to provide the diversity in IC ordering necessary to enable LifeGPT to capture all of the rules of Life accurately. This version of LifeGPT effectively found a `pocket of computational reducibility'~\cite{wolfram_can_2024} within Life which applied only to ICs outside of a particular order parameter range.

The second conclusion is that this state-transition asymmetry problem can be effectively remedied by increasing the breadth of \(\langle\eta\rangle\) for samples in training data. The broad-entropy set, containing a relatively flat distribution of IC ordering (Fig. \ref{fig:order_param_dist}B), was capable of providing the sample diversity necessary for the model to learn Life's rules with near perfect accuracy.

\subsection{Zero/Few Shot Learning Abilities}\label{sec:train_set_size}
Zero-shot and few-shot learning are demonstrated when models perform well on tasks that either do not appear or rarely appear in training data. We show that our training data, containing 10,000 stochastically-generated ICs (see \ref{sec:testing_set_gen}), represents only a small fraction of all possible ICs. It is extremely improbable for one or more of the ICs in our testing set (excluding the two samples generated with \(p_x=0\) and \(p_x=1\) respectively) to exactly match one or more ICs in our training set. Given that the total number of possible initial conditions is \(N_{\mathrm{Total}} = 2^{1024} \approx 1.80 \times 10^{308}\), the probability of randomly testing the model on a training sample is \(\frac{10,000}{2^{1024}} \approx 5.56 \times 10^{-305}\). We also empirically confirmed that there were no matches between the 8 testing sample ICs (again, excluding $p_x=0$ and $p_x=1$) and the 10,000 sample training set we used, using a simple search algorithm.

\begin{figure}[H]
    \centering
    \includegraphics[width=\columnwidth]{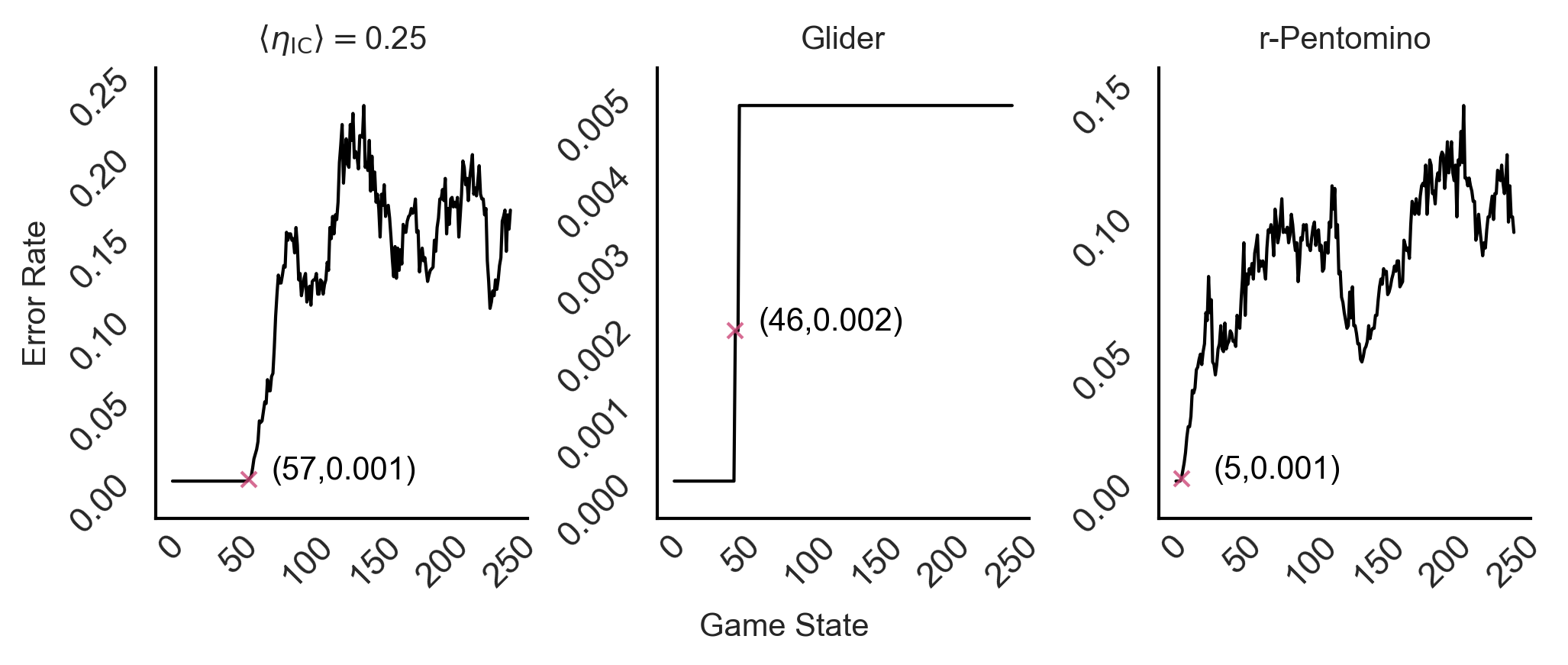}
    \caption{Error rate vs. game state for 3 samples in the testing set for which ARAR was run and compared against the GT Life algorithm. Red `X's mark the 1st game states where the error rate became greater than 0.}
    \label{fig:3_samples_error}
\end{figure}

Based on our performance characterization (Fig. \ref{fig:Model_Performance}), LifeGPT displays impressive zero/few-shot learning capabilities. LifeGPT is able to simulate the rules of Life with near-perfect accuracy for a 32 $\times$ 32 toroidal grid by learning off of less than one part in one cubic-googol (i.e. $<10^{-300}$) of all possible starting configurations. Still, extremely rare instances of incorrect token outputs are observed, even with a sampling temperature of 0. This is observed experimentally when feeding the output of LifeGPT back into its input, creating two layers of autoregression (we called this method ARAR, for `autoregressive autoregressor' -- see section \ref{sec:ARAR}), which recursively implement the rules of Life. Specifically, an error in a the output of a single cell will cause a perturbation to the Life system, that, over several additional iterations, is highly likely to drastically affect the entire game state with respect to the ground truth. This occurred when inputting the `Order\_param: 0.25' (\(\langle\eta\rangle_{\mathrm{IC}}=0.25\)), `glider,' and `r-Pentomino' ICs from the testing into the ARAR algorithm (see \href{https://github.com/lamm-mit/LifeGPT/blob/main/Testing_Set_ARAR_Animations_and_Figures/Animations/Epoch_50_Temp_0_Timesteps_250/evolution_temp_0_sample_2.gif}{\(\langle\eta\rangle_{\mathrm{IC}}=0.25\) Animation at Temperature 0 -- 250 Timesteps}, \href{https://github.com/lamm-mit/LifeGPT/blob/main/Testing_Set_ARAR_Animations_and_Figures/Animations/Epoch_50_Temp_0_Timesteps_250/evolution_temp_0_sample_6.gif}{Glider Animation at Temperature 0 -- 250 Timesteps}, and \href{https://github.com/lamm-mit/LifeGPT/blob/main/Testing_Set_ARAR_Animations_and_Figures/Animations/Epoch_50_Temp_0_Timesteps_250/evolution_temp_0_sample_10.gif}{r-Pentomino Animation at Temperature 0 -- 250 Timesteps}, respectively). For the \(\langle\eta\rangle_{\mathrm{IC}}=0.25\) and `r-Pentomino' ICs, only a single cell was incorrectly predicted during the generation step which caused divergence from ground truth, while for the `glider' IC, divergence was observed due to an iteration wich incorrectly predicted 2 cells (Fig. \ref{fig:3_samples_error}).

\begin{figure}[H]
    \centering
    \includegraphics[width=\columnwidth]{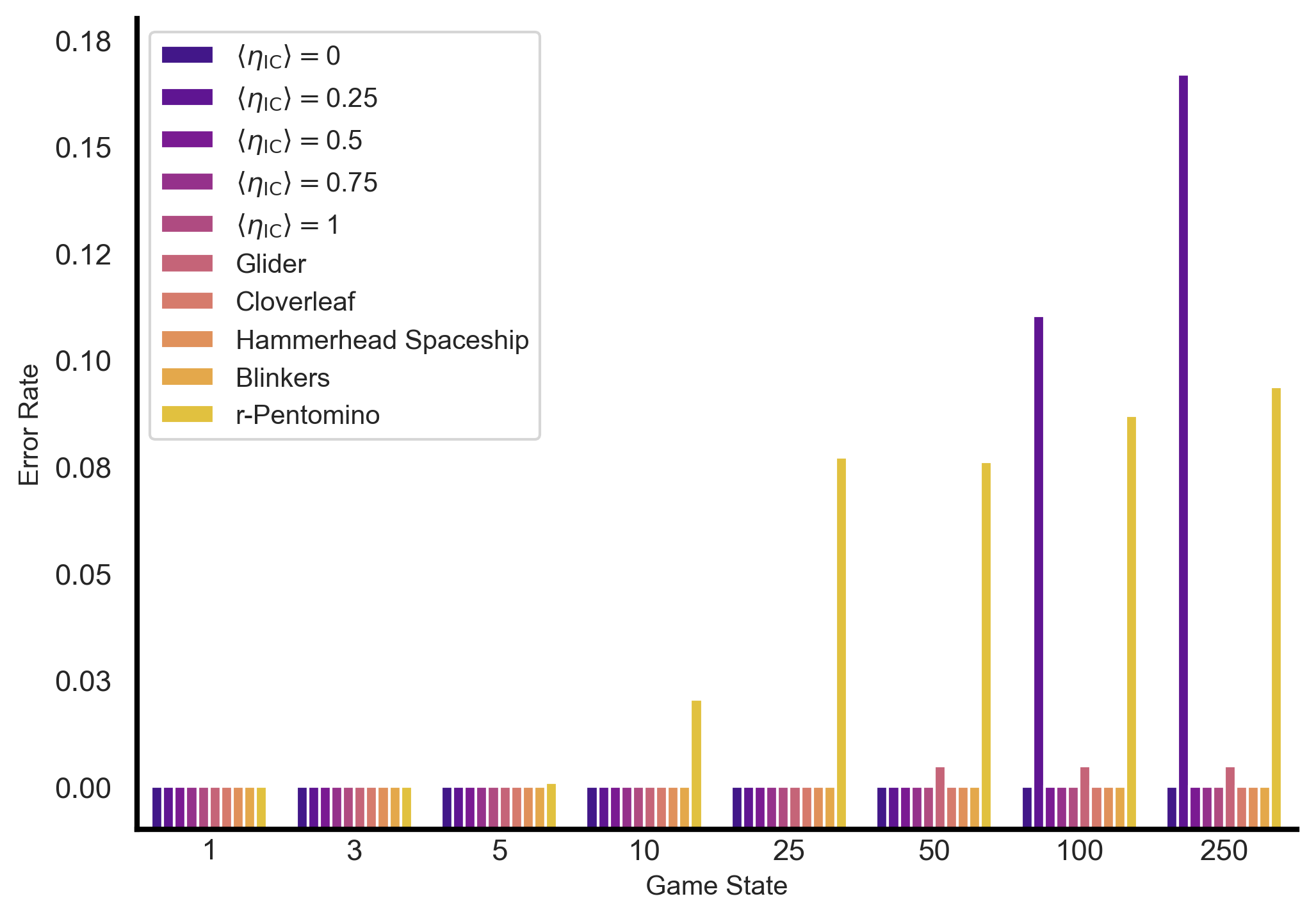}
    \caption{A histogram showing the fraction of cells that were incorrectly predicted by LifeGPT (using ARAR) with respect to GT, for varying future game states (where the IC is the 1st game state), for varying samples in the testing set. All but 3 samples show no error rate above 0 for all game states.}
    \label{fig:histo}
\end{figure}

These were the only instances of inaccurate predictions made by LifeGPT (the best-performing version trained for 50 epochs on the broad-entropy training set) with the temperature set to 0 that we encountered during all of our testing (Fig.~\ref{fig:histo}).

Extended Data Fig.s \ref{fig:sample_1}, \ref{fig:sample_2}, \ref{fig:sample_3}, \ref{fig:sample_4}, \ref{fig:sample_5}, \ref{fig:sample_6}, \ref{fig:sample_7}, \ref{fig:sample_8}, \ref{fig:sample_9}, and \ref{fig:sample_10} show several iterations of the ARAR algorithm using LifeGPT to predict NGSs, with respect to ground truth (GT) Life iterations, starting from the same ICs in the testing set. LifeGPT's error with respect to GT is also depicted as a binary grid for each game state shown. Extended Data Fig.~\ref{fig:sample_10} clearly shows that, for the 5th game state (where the IC for 'r-Pentomino' is the 1st game state), LifeGPT predicts a single incorrect token. This initial error was shown to create a feedback loop of growing divergence between LifeGPT's prediction and the GT Life evolution. 

Importantly, this suggests that LifeGPT is not, in fact, exactly replicating the Life algorithm, but is instead approximating it in an extremely effective, while occasionally fallible, manner. Notably, the patterns for which LifeGPT exhibited this failure tended to be ones where most of the grid was full of 0-state cells, aside from small clusters of 1-state cells near the edges of the grid (see Extended Data Fig.s~\ref{fig:sample_2}, \ref{fig:sample_6}, and \ref{fig:sample_10}), suggesting that the heterogeneity of training data samples also affects converged accuracy, and possibly that the model struggled to understand the toroidal grid topology (periodic grid boundary conditions) for \(\eta_{\mathrm{IC}}\gtrsim0\). It is possible that increasing heterogeneity of the entropies of the ICs -- including ICs where different subsections of each grid exhibited vastly different ordering -- in the training data might have a positive effect for these `edge-cases.' Alternatively, it is possible that the LifeGPT would have not made this error had it been trained for substantially more than 50 epochs, although this is unlikely considering CEL showed few indications of an overall downward trend by epoch 50 (Fig.~\ref{fig:Model_Performance}A).

\subsection{Future Work}
LifeGPT's ever-increasing likelihood of generating incorrect tokens with increasing temperature suggests a fundamental trade-off between model creativity and accuracy in making deterministic predictions. This key result suggests that LifeGPT learns an approximation of the rules that fails under specific circumstances. This may be due to an intrinsic limitation of the transformer architecture for modelling deterministic systems; nevertheless a future version of LifeGPT could be used to simulate CA systems which do not require 100\% accurate predictions, or for which `perfect' predictions are impossible, such as in the case of stochastic CA, or modeling real-world mechanisms such as growth or dynamics of living cells, amongst others. Another possible avenue of exploration would be to use GPT models to predict coarse-grained CA evolution, as previous work has shown that even highly complex CA often have simpler course-grained representations using a renormalization group approach~\cite{israeli_coarse-graining_2006}. 

Future LifeGPT-like models, trained on a massive space of CA rulesets (akin to a tiny piece of the `ruliad'~\cite{wolfram_concept_2021}), may be able to solve a multitude of inverse problems. Here, instead of predicting the precise evolution of CA systems, they may be capable of assigning a representative CA ruleset to a real-world dynamical system to raw data representing the system's spatiotemporal evolution. This rules-based representation would help advanced the phenomenological understanding of the system at hand, as different coarse-grained rules could be uncovered for different length-scales, and would enable generative exploration of the system's configuration-space via IC-variation and subsequent recursive rule application. For instance, this strategy could be applied to better understand the growth-dynamics of plants and fungi, since CA have already been tried as models for many such systems~\cite{komarov_concept_2003,balzter_cellular_1998,halley_competition_1994,ferreira_modelling_2013,laszlo_cellular_1993}, and the growth of such organisms is largely governed by the emergent effects of cellular interactions~\cite{dobrescu_emergence_2011,jakab_engineering_2004,xiao_dna_2018,newman_before_2006}. This concept echoes the existing idea that AIs might be able to find ``pockets of computational reducibility'' in otherwise computationally irreducible systems.Some theorize that transformer models could prune search spaces in multicomputational systems, where non-deterministic state transitions explore multiple paths to reach final states. This raises the question of whether AIs could similarly find rule sets that meet desired emergent behaviors, with the answer depending on the structure of the rule space. One hypothesis on this subject suggests that higher-dimensional rule spaces may allow enough emergent complexity for random rule changes to effectively escape local minima on cost surfaces~\cite{wolfram_can_2024}.

Yet another opportunity may arise from the fact that Life has been shown to be Turing complete, via the implementation of a Universal Turing Machine (UTM)~\cite{rendell_game_2016}. Turing completeness is, in most cases, synonymous with computational universality\cite{delvenne_computational_2005}. If a system can be proven to be Turing complete, then that a system is capable of simulating any Turing machine, or in other words, any classical computer, granted the system is large enough to store the necessary code~\cite{reprintsev_turing_2018}. Therefore, our work empirically shows that, theoretically, GPT models could become capable of simulating any classical computer algorithm (granted sufficient hardware). This falls in line with theoretical work that has proven the Turing-completeness/computational universality of the transformer architecture ~\cite{perez_turing_2019, perez_attention_2021} as well as autoregressive LLMs specifically~\cite{schuurmans_autoregressive_2024}. While our GPT model was trained on game grids far too small to implement a full-scale UTM within Life, and modern-day hardware still limits the practicality of synthesizing a GPT with Life to simulate a UTM, our work nevertheless showcases the possibility for future models to synthesize stochastic generative capabilities with deterministic computational capabilities. This would be particularly useful in the context of hallucination-prone LLM chatbots~\cite{bhattacharya_strategies_2024,hamid_beyond_2024}, as simulating CA-based computers might enable them to perform deterministic calculations without compromising their creative, generative capabilities.

Considering that LifeGPT does, occasionally, predict incorrect tokens, more work could be done to investigate other methods of training GPTs on this front. In recent years, researchers have investigated the potential of transformer models -- particularly LLMs -- to excel at strategic game-play~\cite{hu_survey_2024}. In particular, methods which incorporate some kind of reinforcement learning (RL), which involves the LLM making decisions within and getting feedback from a simulated game environment, have been shown to be particularly successful. For these types of applications, transformer models -- specifically, LLMs -- act as agents which can make a series of decisions, often producing more than one output at a time, conveying several viable options. An RL algorithm works in tandem with the LLM to choose the best decisions from generated options. By playing a given strategy game repeatedly against other AI agents, where each decision leads to a generated reward enabling RL training, the agents will learn over time to take optimal decisions. This strategy usually requires a pretrained LLM chatbot, and does not necessarily involve any LLM fine-tuning steps. Xu \textit{et al.} (2024) found that the addition of an RL algorithm, as opposed to only using an LLM, helps the model overcome intrinsic biases found in LLMs such as ChatGPT. For a so-called `solved game,' this strategy can ensure that the model's actions more closely align with the corresponding Nash equilibrium~\cite{xu_language_2024}.

When learning Life, the situation is slightly different, as Life is a `0 player game,' meaning it behaves more like a simulation, where a user simply inputs some IC and lets the program run its course. Therefore, a language-agent-plus-RL approach for predicting Life dynamics would rely on creating a `meta-game,' a game in which the Nash equilibrium results in the duplication of Life's rules. We believe such a strategy is worth exploring in future work, as it may serve to mitigate errors. We further envision that a GPT pretrained accurately on a number of diverse CAs can be a powerful foundation model to learn CA rules from real biological data, for instance, for which exact algorithms may not exist. It could potentially also serve as a way to solve the inverse problem, to identify an algorithm for a given set of data. 

Going further, Liu \textit{et al.} (2024) reconceptualizes the LLM learning processes in the context of agent learning in two-player language games~\cite{liu_large_2024}. They go on to use their new framework to argue that hallucinations in current, state-of-the-art LLMs is due to a lack of a `world model'~\cite{ha_world_2018}. In the context of a model trained to predict the next state in a CA algorithm such as Life, a world model could be represented as a state-transition table or graph. Future LifeGPT-like models could therefore benefit from being tasked with, in addition to predicting NGSs, predicting state-transition rulesets. Hypothetically, if the training data were then to be structured to incorporate many examples of CA ICs and NGSs, along with corresponding rulesets, the model could learn to interpret the ruleset as a world model, allowing it to check its NGS outputs to mitigate errors. Such an approach would be uniquely interesting for the aforementioned task of assigning rulesets to real-world biological systems, as the world model could quadruply serve as a error-mitigating component of the main GPT model, a model interpretability tool, a biological system interpretability tool, and a generative design tool.

\section{Conclusion}
We have demonstrated the functionality of LifeGPT, a decoder-only GPT model trained on NGS transition examples from Conway's Game of Life. By allowing LifeGPT to learn the dynamics of Life, we demonstrated, to the best of our knowledge, the first ever transformer model that has been trained to predict NGS transitions in a 2D CA system.  

LifeGPT represents a step forward in the synthesis of machine learning with ALife, and in the understanding of highly complex systems with emergent properties. The demonstrated capability of LifeGPT (which is orders of magnitude smaller than SOTA GPT models) to learn the state-transition rules governing Life with near-perfect accuracy opens up new opportunities for both understanding and modelling highly systems highly sensitive to initial conditions, as well as universal computation in an LLM framework, specifically pretrained LLMs, that merges the capacity to model CAs with a host of other capabilities. 

Future work could also focus on mitigating the effects of occasional errors. The intrinsic stochasticity of the transformer architecture leaves the door open for building better understandings of probabilistic systems, and of coarse-grained approximations of the real-world. The topology-agnostic nature of LifeGPT, unlike previous methods of learning Life and other CA, may allow future exploration of a diverse space of nonlinear, dynamical, continuum systems, regardless of the details of spatiotemporal dependencies.

\section{Materials and Methods}\label{sec:methods}
Codes, data, and relevant animations/figures are available at \href{https://github.com/lamm-mit/LifeGPT}{https://github.com/lamm-mit/LifeGPT}.

\subsection{Model Architecture and Hardware Information}
LifeGPT was constructed in python using the `x-transformers' library\cite{wang_lucidrainsx-transformers_2024}. The models in this study were trained with a workstation equipped with a high-end CUDA-compatible GPU (RTX A4000, NVidia, Santa Clara, CA, USA) for a total of 50 epochs on a 10,000-sample training set.

\subsection{Hyperparameters}\label{sec:hyper}
Hyperparameters were initially selected heuristically for optimal performance, as the GPU primarily used for training (RTX A4000, NVidia, Santa Clara, CA, USA) had 16 GB of VRAM. Unless otherwise stated, all instances of LifeGPT used the following set of hyperparameters during training, as described in Table~\ref{tab:hyper}. The batch size was initially set to 20 samples, and was decreased to 5 samples for later versions of LifeGPT due to memory limitations encountered when using FCM (see section \ref{sec:FCM}).
\begin{table}[!htbp]
  \caption{LifeGPT's hyperparameters (best-performing).}
  \centering
  \begin{tabular}{ll}
    \toprule
    \textbf{Hyperparameter}      & \textbf{Value}         \\
    \midrule
    num\_tokens             & 256                    \\
    max\_seq\_len           & 2071                   \\
    dim (\(d_{\mathrm{model}}=d_{\mathrm{embedding}}\))                     & 256                    \\
    depth (\(N_{\mathrm{layers}}\))                   & 12                     \\
    heads (\(h\))                   & 8                      \\
    attn\_dim\_head (\(d_{k}\))         & 64                     \\
    RPE        & \texttt{True}          \\
    Flash attention             & \texttt{True}          \\
    Optimizer               & Adam
    \\
    Learning rate          & 1e-4
    \\
    mask\_prob              & 0.15
    \\
    Loss function           & Cross entropy loss (CLE)
    \\
    Gradient accumulation period    & 5
    \\
    Training data IC ordering & broad-entropy
    \\
    \bottomrule
    \label{tab:hyper}
  \end{tabular}
\end{table}

\subsection{Datasets}
\subsubsection{Data Generation Overview}
To generate training sets, validation sets, and testing sets, the same basic strategy was used. First an IC game-states would be generated stochastically as a 2D, 32 $\times$ 32 numpy array. Depending on the exact algorithm used, the generated IC game-states would collectively form either high-entropy or broad-entropy datasets. Next, a proprietary Life python class was used to generate the corresponding NGS for every previously generated IC. Lastly, each IC and corresponding NGS was concatenated within a string. Every generated pair was subsequently stored within a dataframe from future retrieval.

\subsubsection{Data Topology.}
Transformer models are architected to process data as 1D arrays. Therefore, to teach the LifeGPT the rules of a 2D CA algorithm, such as Life, the 2D data from each time-slice of the game had to be flattened into a 1D array. In this way, LifeGPT functioned similar to a vision transformer, in which 2D data is flattened into a 1D array within which each entry is a tokenizable image patch~\cite{berroukham_vision_2023}. However, due to the low-resolution of the 32 $\times$ 32 toroidal grid on which Life was simulated to generate our training, we were able to encode every pixel of each time-slice of the game in a 1D array (as opposed to grouping pixels into patches). 

\subsubsection{Instruction Tuning.}

In order to encode the time-progression of the game into the training set, the initial-state and next-state 1D arrays were placed within a prompt-string which was subsequently tokenized to form a vector. Specifically, both 1D arrays were converted to strings and placed within a larger string containing start and end tokens (\verb+@+ and \verb+$+, respectively), a task statement, and bracket delimitors (e.g. `\verb+@PredictNextState<INITIAL_CONDITION> [NEXT_STATE]$+'). 

\subsubsection{Tokenization.}

We employed a byte-level tokenizer that operates on UTF-8 encoded text. UTF-8 is a variable-width character encoding capable of representing every character in the Unicode standard, which allows the tokenizer to process a wide range of scripts, symbols, and special characters uniformly. By converting the text into its byte-level representation, our approach ensures consistent tokenization across different languages and handles out-of-vocabulary words and non-standard text, such as emojis or code, effectively. This method allows for robust and flexible processing of diverse textual data. Tokenization resulted in a vector suitable as input to the embedding layer of the transformer model. 

\subsubsection{Training Set Generation}\label{sec:train_set_gen}
\paragraph{High-Entropy IC Set Generation}\label{sec:hientgen}
High entropy IC game-states were generated by effectively flipping a coin 1024 times to designate the states (0 or 1) on a 32 $\times$ 32 grid. When considering the configuration space of a binary 2D array \( M \in \{0, 1\}^{32 \times 32} \), the following formula may be used to describe its Shannon entropy~\cite{shannon_mathematical_1948} (informational entropy):

\begin{equation}
H(M) = -\sum_{x \in \{0, 1\}} p_x \log_{2} p_x \label{eq:entropy}
\end{equation}

(this is also known as the binary entropy function~\cite{mackay_information_2003}) where, \( p_x \) is the probability of finding the value \( x \) in the \(32 \times 32\) array \( M \). \( p_x \) is defined as:

\begin{equation}
p_x = \frac{1}{32^2} \sum_{i=1}^{32} \sum_{j=1}^{32} \delta_{M_{ij}, x} \label{eq:px}
\end{equation}

where, \(M_{ij}\) is an element of $M$ in the $i$th row and $j$th column, and \( \delta_{M_{ij}, x} \) is the Kronecker delta function, which is equal to 1 if \( M_{ij} = x \) and 0 otherwise.

Thus, for a `50-50 coin toss' scenario (\( p_0 = p_1 = \frac{1}{2} \)), \( H(M) \) is at a maximum and is equal to 1 Sh. Moreover, since binary data necessitates the condition \( p_0 + p_1 = 1 \), only one probability value is needed to fully describe the entropy of a given array \( A \). We therefore denote the ordering of a given IC by referring to a single order parameter, \(\eta\), where \(\eta = p_1\) is always true. When considering the order parameter of a set of ICs, it is important to note that, because IC generation is always a stochastic process, the exact \(\eta\) of any given IC in the set cannot be predicted with certainty. For this reason, we characterize IC sets with the symbol \(\langle\eta\rangle\), denoting the expected order parameter.

To generate high-entropy ICs, a binary array was constructed by checking \verb+random.random() < 0.5 == True+ (using the `random' module in python -- see \href{https://python.readthedocs.io/en/latest/library/random.html}{https://python.readthedocs.io/en/latest/library/random.html}) to decide each element. If the statement returned true, then the element would be defined as 1, and otherwise, 0. This method resulted in a training set with a binomial, experimentally measured \(\eta\) distribution (Fig.~\ref{fig:order_param_dist}A)

\paragraph{Broad-Entropy IC Set Generation}
\label{sec:broadentgen}

To create a broad-entropy IC set, first, a vector was created representing a set of order parameters ranging from 0 to 1. The length of this vector was set to the desired number of samples in the dataset (10,000 for training, 1000 for validation). This set of order parameters may be thought of as containing different expected probabilities for finding a 1 in an IC.

Then, the same procedure as with the high-entropy IC set was followed, with two exceptions: (1) instead of \verb+random.random() < 0.5 == True+ determining the value of each element in each IC array, \texttt{random.random() < $\eta$ == True}
 was the determining equality, and (2) each IC was generated using a unique \(\eta\) from the aforementioned vector (see section \ref{sec:hientgen}). This strategy ensured that the IC set represented a broad range of ordering, from all 0s, to 50-50 0s and 1s, to all 1s (Fig.~\ref{fig:order_param_dist}B).

\paragraph{Next-Game-State Generation.}
\label{sec:nextstategen}
NGSs were calculated from IC arrays by applying Life rules assuming a toroidal grid (see the \verb+update_grid()+ function here: \href{https://github.com/lamm-mit/LifeGPT/blob/main/conway_lib/game.py}{game.py}). 

\paragraph{Reshaping Data.}
\label{sec:reshaping}
To make the handling of training set data easier, the final stage of the training set generator involves reshaping the data into a list of sub-lists, in which each entry in the list contains a sub-list corresponding to a specific IC. Within each unique sub-list, two strings are stored, one corresponding to a flattened IC, and one corresponding to a flattened NGS (see the \verb+generate_sets()+ function here: \href{https://github.com/lamm-mit/LifeGPT/blob/main/conway_lib/game.py}{game.py}). 

\subsubsection{Validation Set Generation}
Validation sets were generated using the same methods in section \ref{sec:train_set_gen}, as the \verb+random.random()+ function ensures sufficiently random IC generation, ensuring training and validation sets remained entirely independent. Combined with the incredibly large space of possible 32 $\times$ 32 binary arrays ($2^{32\times32}\approx 1.80\times 10^{308}$ unique possibilities), this made the likelihood of even a single sample being identical between a 10,000-sample training set and a 1000-sample validation set negligible (see section \ref{sec:train_set_size}). This, in turn, ensured that over the course of model training, training loss and validation loss remained independent of one another. 

\subsubsection{Testing Set Generation}
\label{sec:testing_set_gen}
A 10-sample testing set was constructed to validate the performance of models during and after training, in a manner other than by inspecting the validation and training losses. 5 samples in the testing set were generated stochastically in the same manner as in section \ref{sec:broadentgen}, and 5 samples were manually defined to match known periodic and complex patterns found in Life (Fig.~\ref{fig:testing_set}). NGSs were recursively generated for a total of 10 states (including the IC) per sample, for all 10 samples in the testing set.

\subsection{Forgetful Causal Masking (FCM) Implementation}
\label{sec:FCM}
FCM was implemented using the `x-transformers' library~\cite{wang_lucidrainsx-transformers_2024}. FCM was built into this library as part of the \texttt{AutoregressiveWrapper} class by default. FCM was enabled by setting \texttt{mask\_prob} to 0.15, which was shown to be effective for zero and few-shot learning by Liu \textit{et al.}~\cite{liu_towards_2023}.

\label{sec:maskingeffects}
FCM involves randomly masking a predetermined percentage of  past-tokens during the learning process, in addition to standard causal attention masking. The authors~\cite{liu_towards_2023} argue that this method prevents over-attending to more recent tokens in a given sequence, encouraging attention to tokens in the `distant past.' We implemented FCM into our model, which increased the rate at which model accuracy improved with each epoch. Furthermore, FCM enabled our model to achieve 100\% accuracy  on our testing set with a temperature of 1.0 in less than 50 epochs, which was previously unattainable when training with a broad-entropy dataset.

Implementing FCM increased the GPU RAM requirements of our LifeGPT, necessitating a decrease in batch size from 20 to 5 samples. 

\subsection{Model Development}
Training was initially conducted with high-entropy data. Due to the (pseudo)random nature of our training set generation script (see Section \ref{sec:train_set_gen}), and the high number of samples in the training set (10,000), there was some diversity of training data entropy despite  use of a static order parameter of ($\eta$=0.5) (Fig.~\ref{fig:order_param_dist}A). Nevertheless, observed model accuracy issues for low-entropy ICs prompted the use of broad-entropy datasets (Fig.~\ref{fig:order_param_dist}B), which resulted in for improved performance. 

\subsection{Accuracy Benchmarking}
\label{sec:model_train}
The testing dataset consisted of 10 flattened 32 $\times$ 32 binary arrays, representing initial states in Life, and their resulting iterations in accordance with Life state-transition rules on a toroidal (periodic) grid, numbering one through ten. Depending on the type of model being trained (the number of desired time-step jump predictions), different columns in the testing dataset would be selected as the ground truth. Accuracy at each checkpoint (every 2 epochs, starting with epoch 2) was determined by inputting the task statement (e.g. \verb+@PredictNextState<INITIAL_CONDITION>+) into a tokenizer, and subsequently using the tokenized data as the prompt for the autoregressive model. Since all of LifeGPT's training was conducted on data corresponding to a 32 $\times$ 32 grid, LifeGPT was programmed to output the exact number of tokens necessary to fully describe the NGS. After LifeGPT was finished generating the output data, this data was compared to the ground truth (the flattened NGS in accordance with Life's rules), and an accuracy score was computed using the following function:

\begin{equation}
\label{eq:acc}
A = \frac{1}{N} \sum_{i=1}^{N} \delta_{y_i \hat{y}_i}
\end{equation}

where \(A\) is the Accuracy of the model, \(N\) is the total number of cell predictions across the testing dataset ($N=32\times32\times10=10240$ cells for a dataset with ten pairs of 32 $\times$ 32 grid game examples), \(y_i\) is the ground truth value, \(\hat{y}_i\) is the predicted value, and \(\delta\) is the Kronecker delta function which equals 1 if \(y_i = \hat{y}_i\) and 0 otherwise. An accuracy score was computed once every 2 epochs, for each model temperature in the set {0, 0.25, 0.5, 0.75, 1}, starting with epoch 2.

\subsection{Training Set Entropy Effects Experimental Procedure}
The goal of this experiment what to determine was effect, if any, the ordering of the ICs making up the training data for LifeGPT would have on accuracy (\(A\)), when fed ICs generated with varying expected order parameters (\(\langle\eta\rangle_{\mathrm{IC}}\)). We used two versions of LifeGPT; one was trained on high-entropy training data, and the other on broad-entropy training data. Next, a broad-entropy testing set (comprised of 110 samples, each with a (\(\langle\eta\rangle_{\mathrm{IC}}\)) value ranging linearly from 0 to 1) was generated in the same manner as the broad-entropy training set. The stochasticity of the IC generation process ensured both broad entropy sets remained independent. Finally, both models were benchmarked on each sample in  a manner similar to the method in section \ref{sec:benchmarking}, the only difference being that \(A\) was calculated for each sample in the testing set, as opposed to an average of all samples. Finally, \(A\) versus (\(\langle\eta\rangle_{\mathrm{IC}}\)) was plotted for both models (see Fig.~\ref{fig:acc_ranging_order_param}). 

\subsection{Autoregressive Autoregressor Implementation}\label{sec:ARAR}
ARAR is simply an implementation of LifeGPT where the model is placed inside a loop, where a portion of its output, corresponding to the NGS, is converted into an input tensor and is fed back into LifeGPT, for a desired number of iterations. As such, the NGS outputs of the previous loop iteration serves as the IC in the next loop iteration. In this way, ARAR is able to `run' Life, in a similar recursive manner as the original algorithm. We ran ARAR using two versions of LifeGPT trained on the broad-entropy training set: one which stopped training at epoch 16 (chosen due to this version being the earliest instance of \(A=1.0\)) for temperature$=1$), and one that continued training until epoch 50, across temperatures 0, 0.25 0.5, 0.75, and 1. We compared the NGSs outputted from our ARAR script with the ground truth NGSs, generated with the Life algorithm, and created animations for all model-temperature combinations, showing the progression of the ground truth Life system, the ARAR-generated NGSs, and the discrepancy between the two. 

We also ran the ARAR script (and the Life algorithm) for 249 iterations (resulting in 250 game states, including the ICs), using only the epoch 50, temperature=0 version of LifeGPT due to time and compute constraints, for all 10 samples in the testing set. For each game state, we compared LifeGPT's predictions to the GT Life algorithm's output using the metric `Error Rate', defined as:

\begin{equation}
\label{eq:error_rate}
\mathrm{Error Rate} = 1 - \frac{1}{G} \sum_{i=1}^{G} \delta_{y_i \hat{y}_i}
\end{equation}

where \(\mathrm{Error Rate}\) is the fraction of incorrect cells the model, \(G\) is the total number of cells comprising each game state ($N=32\times32=1024$ cells), \(y_i\) is the ground truth value, \(\hat{y}_i\) is the predicted value, and \(\delta\) is the Kronecker delta function.

\subsection{Use of Generative AI}
Some Python scripts used for data generation, model training, data processing, and figure generation were written with the assistance of GPT-3.5, GPT-4, and GPT-4o from OpenAI. All scripts generated/edited in this manner were carefully reviewed, validated, and manually corrected, in the case of errors, by an author prior to implementation in our work.

\section{Supplementary Information}
Supplementary Information is provided, including a variety of animations and movies. 

\section*{Data Availability}
All data used in the development of this work is available at \url{https://github.com/lamm-mit/LifeGPT}. 

\section*{Code Availability}
All codes are available at \url{https://github.com/lamm-mit/LifeGPT}. 

\section*{Author Contributions}
JAB and MJB conceived the concept, plan of study, developed the model and research, and wrote the paper. JAB developed the algorithms, codes and GitHub repository. JAB conducted the scientific investigations. JAB and MJB wrote, revised and finalized the paper. 

\section*{Conflicts of Interest}
The authors declare no conflict of interest. 

\section*{Acknowledgments}
JAB is supported by the Department of Defense (DoD) through the National Defense Science and Engineering Graduate (NDSEG) Fellowship Program. We acknowledge support from USDA (2021-69012-35978), DOE-SERDP (WP22-S1-3475), ARO (79058LSCSB, W911NF-22-2-0213 and W911NF2120130) as well as the MIT-IBM Watson AI Lab, MIT’s Generative AI Initiative, and Google. Additional support from NIH (U01EB014976 and R01AR077793) is acknowledged.

\section*{Extended Data Figures}

\setcounter{figure}{0} 

\begin{figure}[H]
    \centering
    \includegraphics[width=\columnwidth]{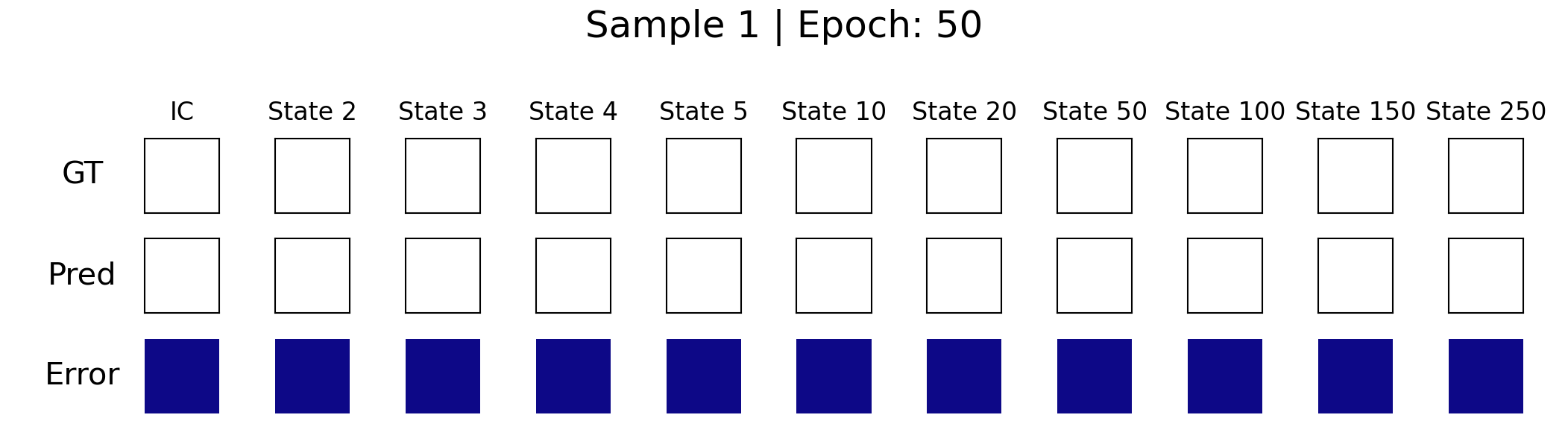}
    \caption{Top row: the ground truth (GT) progression of Life, for an IC corresponding to \(\langle\eta\rangle_{\mathrm{IC}}=0\). Middle row: LifeGPT's best prediction (Pred) using the ARAR method at temperature=0. Bottom row: the discrepancy between the ground truth and LifeGPT's predictions (Error), where blue indicates no discrepancy, and yellow indicates an incorrect cell predicted by LifeGPT.}
    \label{fig:sample_1}
\end{figure}

\begin{figure}[H]
    \centering
    \includegraphics[width=\columnwidth]{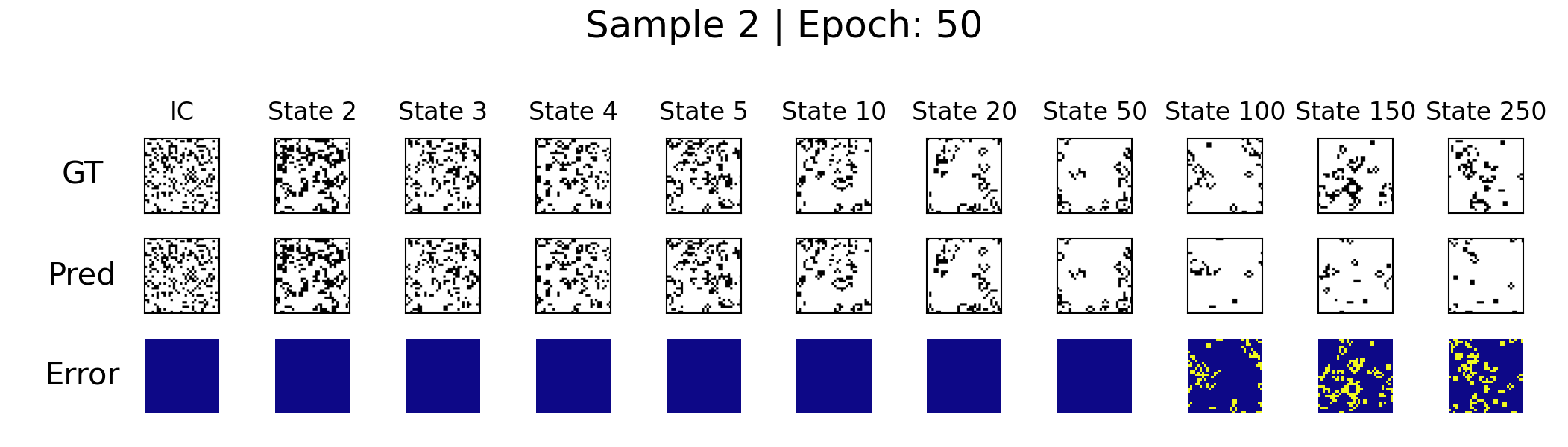}
    \caption{Top row: the ground truth (GT) progression of Life, for an IC corresponding to \(\langle\eta\rangle_{\mathrm{IC}}=0.25\). Middle row: LifeGPT's best prediction (Pred) using the ARAR method at temperature=0. Bottom row: the discrepancy between the ground truth and LifeGPT's predictions (Error), where blue indicates no discrepancy, and yellow indicates an incorrect cell predicted by LifeGPT.}
    \label{fig:sample_2}
\end{figure}

\begin{figure}[H]
    \centering
    \includegraphics[width=\columnwidth]{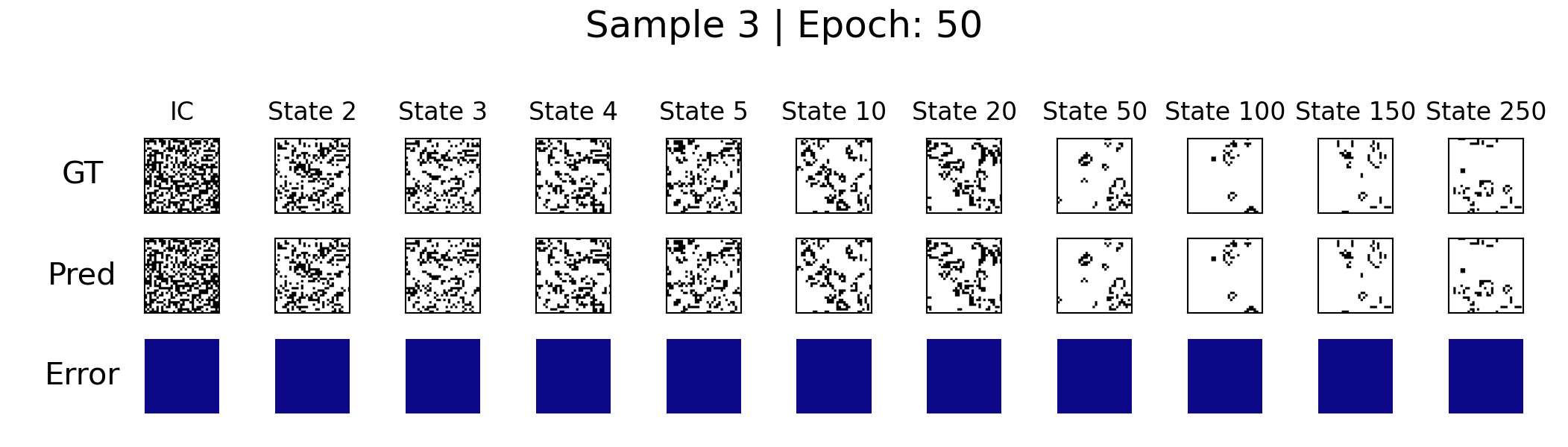}
    \caption{Top row: the ground truth (GT) progression of Life, for an IC corresponding to \(\langle\eta\rangle_{\mathrm{IC}}=0.5\). Middle row: LifeGPT's best prediction (Pred) using the ARAR method at temperature=0. Bottom row: the discrepancy between the ground truth and LifeGPT's predictions (Error), where blue indicates no discrepancy, and yellow indicates an incorrect cell predicted by LifeGPT.}
    \label{fig:sample_3}
\end{figure}

\begin{figure}[H]
    \centering
    \includegraphics[width=\columnwidth]{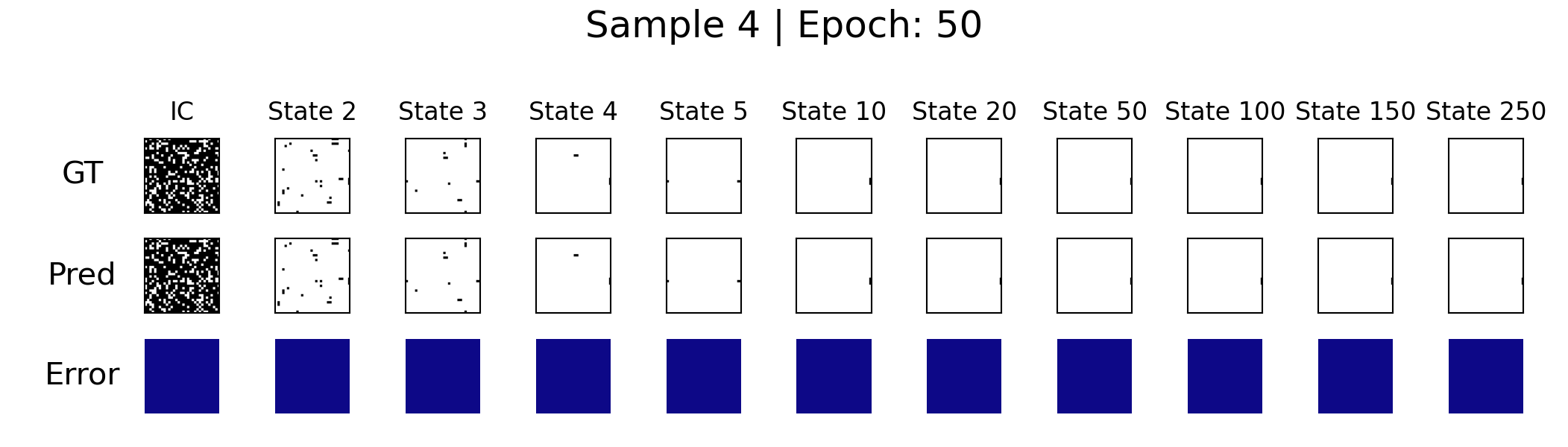}
    \caption{Top row: the ground truth (GT) progression of Life, for an IC corresponding to \(\langle\eta\rangle_{\mathrm{IC}}=0.75\). Middle row: LifeGPT's best prediction (Pred) using the ARAR method at temperature=0. Bottom row: the discrepancy between the ground truth and LifeGPT's predictions (Error), where blue indicates no discrepancy, and yellow indicates an incorrect cell predicted by LifeGPT.}
    \label{fig:sample_4}
\end{figure}

\begin{figure}[H]
    \centering
    \includegraphics[width=\columnwidth]{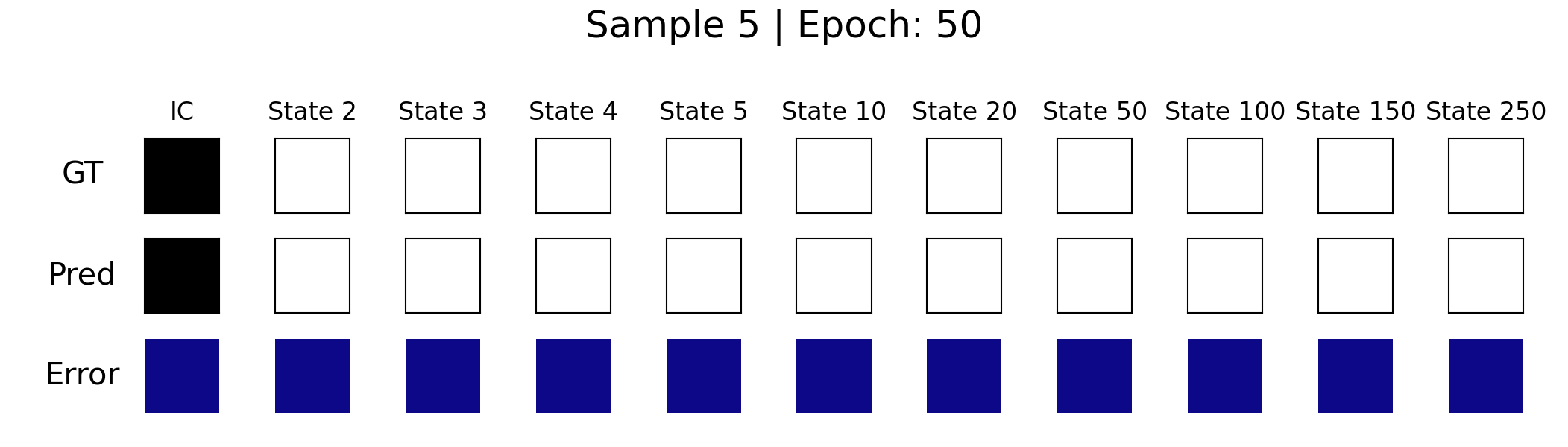}
    \caption{Top row: the ground truth (GT) progression of Life, for an IC corresponding to \(\langle\eta\rangle_{\mathrm{IC}}=1\). Middle row: LifeGPT's best prediction (Pred) using the ARAR method at temperature=0. Bottom row: the discrepancy between the ground truth and LifeGPT's predictions (Error), where blue indicates no discrepancy, and yellow indicates an incorrect cell predicted by LifeGPT.}
    \label{fig:sample_5}
\end{figure}

\begin{figure}[H]
    \centering
    \includegraphics[width=\columnwidth]{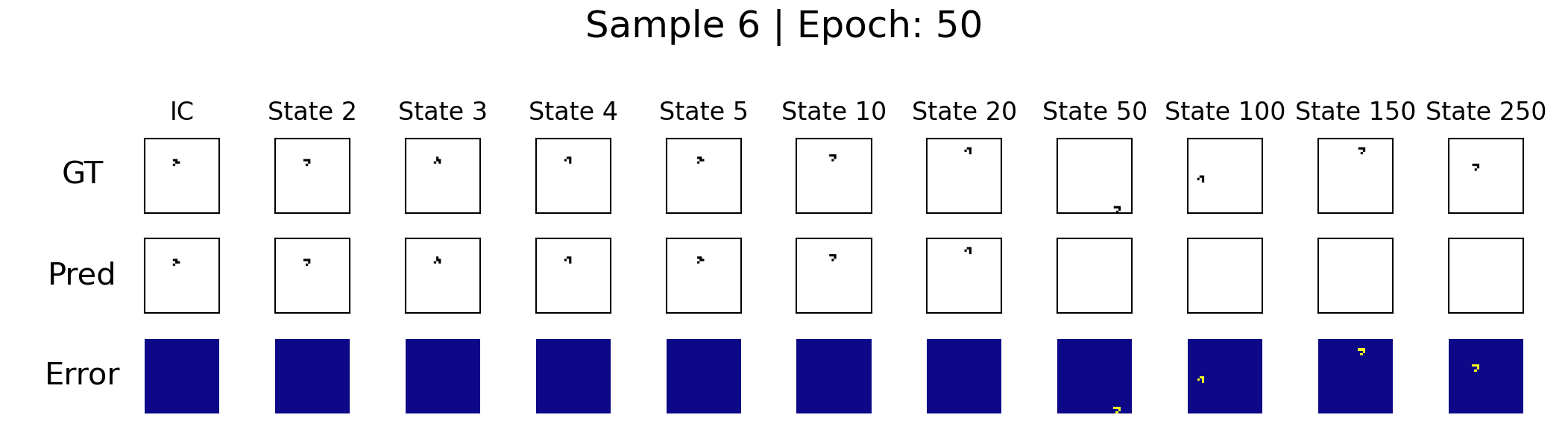}
    \caption{Top row: the ground truth (GT) progression of Life, for an IC corresponding to the `Glider' pattern. Middle row: LifeGPT's best prediction (Pred) using the ARAR method at temperature=0. Bottom row: the discrepancy between the ground truth and LifeGPT's predictions (Error), where blue indicates no discrepancy, and yellow indicates an incorrect cell predicted by LifeGPT.}
    \label{fig:sample_6}
\end{figure}

\begin{figure}[H]
    \centering
    \includegraphics[width=\columnwidth]{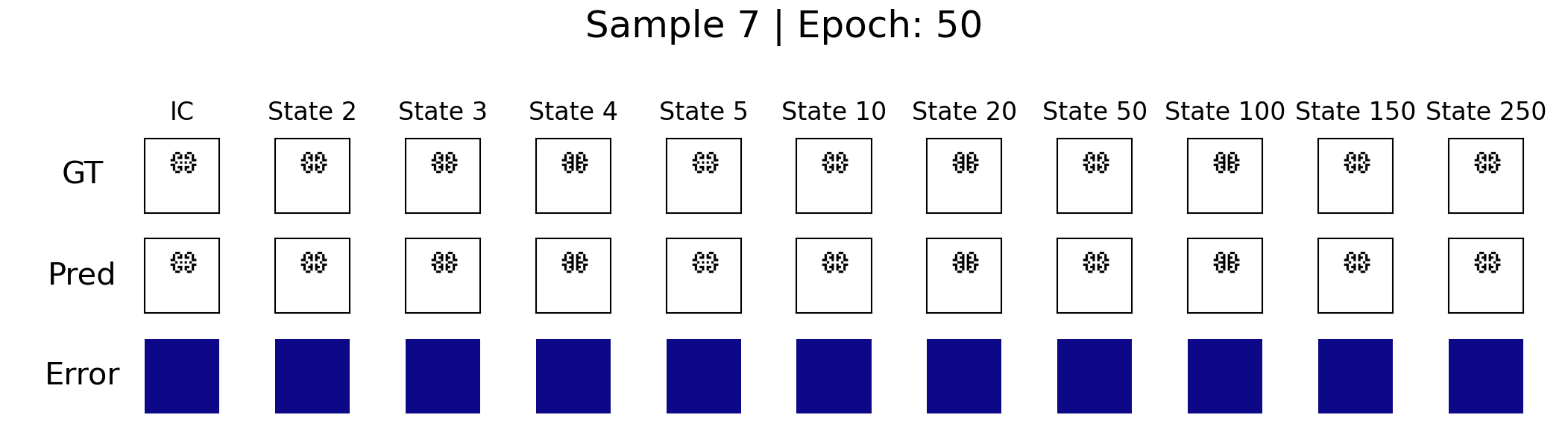}
    \caption{Top row: the ground truth (GT) progression of Life, for an IC corresponding to the `Cloverleaf' pattern. Middle row: LifeGPT's best prediction (Pred) using the ARAR method at temperature=0. Bottom row: the discrepancy between the ground truth and LifeGPT's predictions (Error), where blue indicates no discrepancy, and yellow indicates an incorrect cell predicted by LifeGPT.}
    \label{fig:sample_7}
\end{figure}

\begin{figure}[H]
    \centering
    \includegraphics[width=\columnwidth]{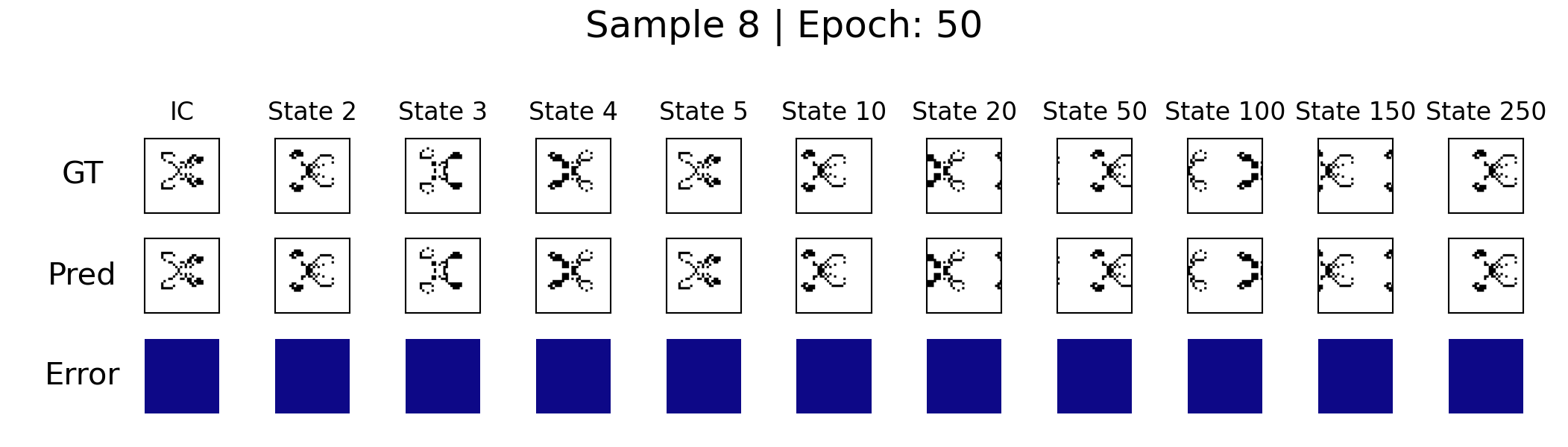}
    \caption{Top row: the ground truth (GT) progression of Life, for an IC corresponding to the `Hammerhead Spaceship' pattern. Middle row: LifeGPT's best prediction (Pred) using the ARAR method at temperature=0. Bottom row: the discrepancy between the ground truth and LifeGPT's predictions (Error), where blue indicates no discrepancy, and yellow indicates an incorrect cell predicted by LifeGPT.}
    \label{fig:sample_8}
\end{figure}

\begin{figure}[H]
    \centering
    \includegraphics[width=\columnwidth]{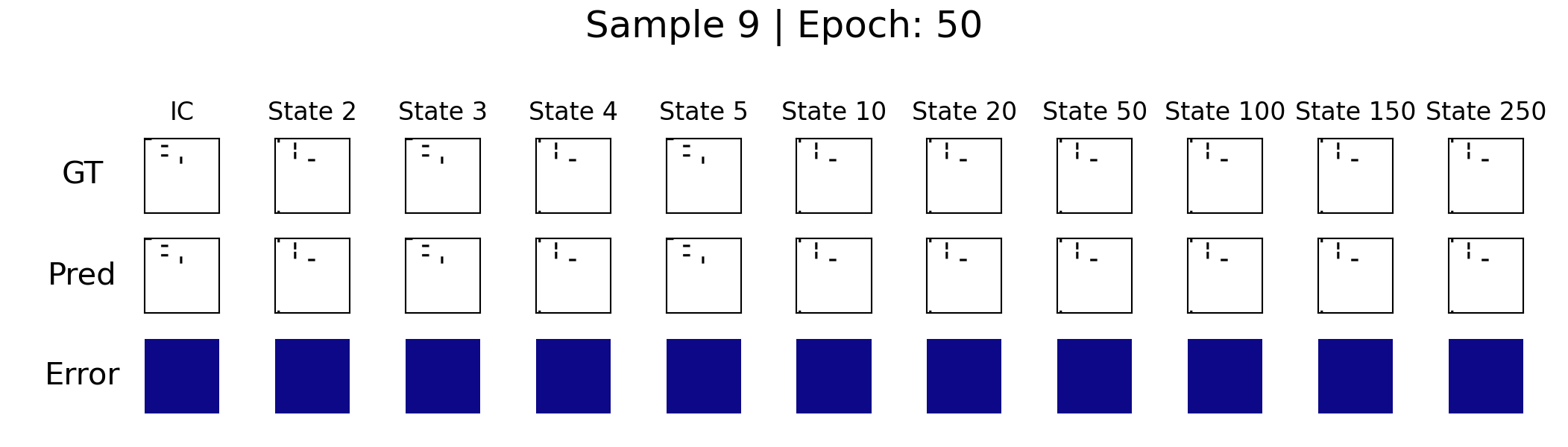}
    \caption{Top row: the ground truth (GT) progression of Life, for an IC corresponding to the `Blinkers' pattern. Middle row: LifeGPT's best prediction (Pred) using the ARAR method at temperature=0. Bottom row: the discrepancy between the ground truth and LifeGPT's predictions (Error), where blue indicates no discrepancy, and yellow indicates an incorrect cell predicted by LifeGPT.}
    \label{fig:sample_9}
\end{figure}

\begin{figure}[H]
    \centering
    \includegraphics[width=\columnwidth]{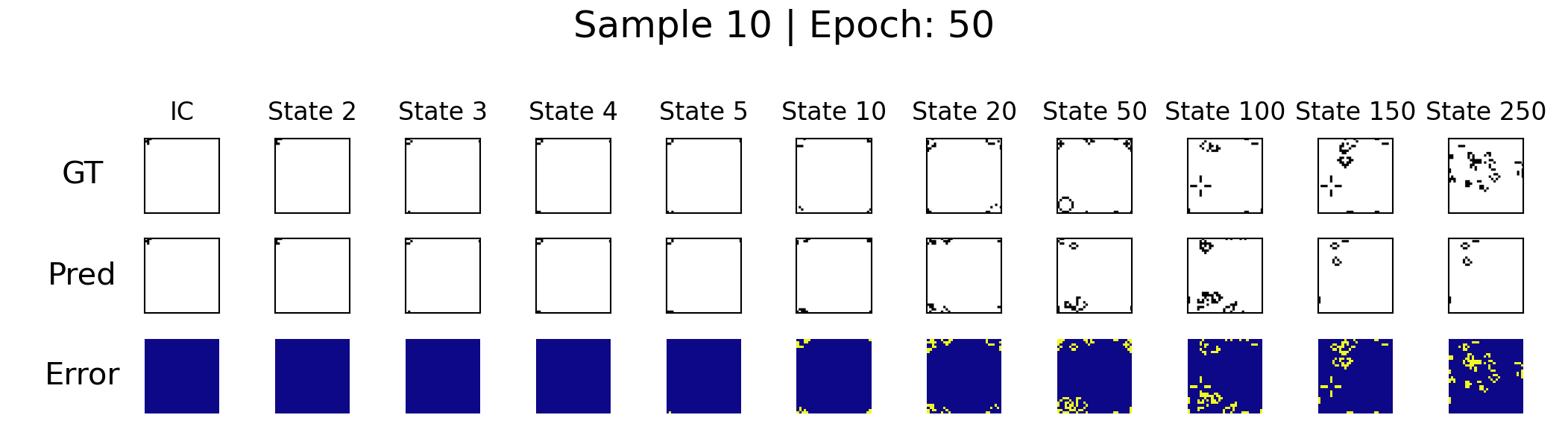}
    \caption{Top row: the ground truth (GT) progression of Life, for an IC corresponding to the `r-Pentomino' pattern. Middle row: LifeGPT's best prediction (Pred) using the ARAR method at temperature=0. Bottom row: the discrepancy between the ground truth and LifeGPT's predictions (Error), where blue indicates no discrepancy, and yellow indicates an incorrect cell predicted by LifeGPT.}
    \label{fig:sample_10}
\end{figure}

\clearpage

\bibliographystyle{unsrt}  
\bibliography{references}

\begin{thebibliography}{10}

\bibitem{hernandez_encinas_simulation_2007}
A.~Hernández~Encinas, L.~Hernández~Encinas, S.~Hoya~White, A.~Martín~del Rey, and G.~Rodríguez~Sánchez.
\newblock Simulation of forest fire fronts using cellular automata.
\newblock {\em Advances in Engineering Software}, 38(6):372--378, June 2007.

\bibitem{zhao_cellular_2020}
Han-Tao Zhao, Xin Zhao, Lu~Jian-cheng, and Liu-yan Xin.
\newblock Cellular automata model for {Urban} {Road} traffic flow {Considering} {Internet} of {Vehicles} and emergency vehicles.
\newblock {\em Journal of Computational Science}, 47:101221, November 2020.

\bibitem{noauthor_scientific_1988}
Scientific {American} {Volume} 259, {Issue} 2, August 1988.

\bibitem{hesselbarth_simulation_1991}
H.~W. Hesselbarth and I.~R. Göbel.
\newblock Simulation of recrystallization by cellular automata.
\newblock {\em Acta Metallurgica et Materialia}, 39(9):2135--2143, September 1991.

\bibitem{wolfram_new_2002}
Stephen Wolfram.
\newblock {\em A {New} {Kind} of {Science}}.
\newblock Wolfram Media, Inc., Champaign, IL, 1st edition edition, May 2002.

\bibitem{gardner_mathematical_1970}
Martin Gardner.
\newblock {MATHEMATICAL} {GAMES}.
\newblock {\em Scientific American}, 223(4):120--123, 1970.

\bibitem{bak_self-organized_1989}
Per Bak, Kan Chen, and Michael Creutz.
\newblock Self-organized criticality in the '{Game} of {Life}".
\newblock {\em Nature}, 342(6251):780--782, December 1989.

\bibitem{wolfram_can_2024}
Stephen Wolfram.
\newblock Can {AI} {Solve} {Science}?, March 2024.

\bibitem{fehsenfeld_persistence_1998}
Kathia~M. Fehsenfeld, M.~A.~F. Gomes, and Tasso R.~M. Sales.
\newblock Persistence of scaling behaviour in the {Game} of {Life}.
\newblock {\em Journal of Physics A: Mathematical and General}, 31(42):8417, October 1998.

\bibitem{martin_inherent_1994}
Bruno Martin.
\newblock Inherent {Generation} of {Fractals} by {Cellular} {Automata}.
\newblock {\em Complex Syst.}, 1994.

\bibitem{wolfram_statistical_1983}
Stephen Wolfram.
\newblock Statistical mechanics of cellular automata.
\newblock {\em Reviews of Modern Physics}, 55(3):601--644, July 1983.

\bibitem{zhang_intelligence_2024}
Shiyang Zhang, Aakash Patel, Syed~A. Rizvi, Nianchen Liu, Sizhuang He, Amin Karbasi, Emanuele Zappala, and David van Dijk.
\newblock Intelligence at the {Edge} of {Chaos}, October 2024.
\newblock arXiv:2410.02536 [cs].

\bibitem{springer_its_2021}
Jacob~M. Springer and Garrett~T. Kenyon.
\newblock It's {Hard} for {Neural} {Networks} to {Learn} the {Game} of {Life}.
\newblock In {\em 2021 {International} {Joint} {Conference} on {Neural} {Networks} ({IJCNN})}, pages 1--8, July 2021.
\newblock ISSN: 2161-4407.

\bibitem{aach_generalization_2021}
Marcel Aach, Jens~Henrik Goebbert, and Jenia Jitsev.
\newblock Generalization over different cellular automata rules learned by a deep feed-forward neural network, November 2021.
\newblock arXiv:2103.14886 [nlin].

\bibitem{weisstein_moore_nodate}
Eric~W. Weisstein.
\newblock Moore {Neighborhood}.
\newblock Publisher: Wolfram Research, Inc.

\bibitem{gray_mathematician_2003}
L.~Gray.
\newblock A {Mathematician} {Looks} at {Wolfram}''s {New} {Kind} of {Science}.
\newblock In {\em Notices {Amer}. {Math}. {Soc}.}, volume~50, 2003.
\newblock Issue 2.

\bibitem{gillioz_overview_2020}
Anthony Gillioz, Jacky Casas, Elena Mugellini, and Omar~Abou Khaled.
\newblock Overview of the {Transformer}-based {Models} for {NLP} {Tasks}.
\newblock In {\em 2020 15th {Conference} on {Computer} {Science} and {Information} {Systems} ({FedCSIS})}, pages 179--183, September 2020.

\bibitem{nguyen_scaling_2023}
Tung Nguyen, Rohan Shah, Hritik Bansal, Troy Arcomano, Sandeep Madireddy, Romit Maulik, Veerabhadra Kotamarthi, Ian Foster, and Aditya Grover.
\newblock Scaling transformer neural networks for skillful and reliable medium-range weather forecasting, December 2023.
\newblock arXiv:2312.03876 [physics].

\bibitem{alerskans_transformer_2022}
Emy Alerskans, Joachim Nyborg, Morten Birk, and Eigil Kaas.
\newblock A transformer neural network for predicting near-surface temperature.
\newblock {\em Meteorological Applications}, 29(5):e2098, 2022.
\newblock \_eprint: https://onlinelibrary.wiley.com/doi/pdf/10.1002/met.2098.

\bibitem{saleem_stc-vit_2024}
Hira Saleem, Flora Salim, and Cormac Purcell.
\newblock {STC}-{ViT}: {Spatio} {Temporal} {Continuous} {Vision} {Transformer} for {Weather} {Forecasting}, May 2024.
\newblock arXiv:2402.17966 [cs].

\bibitem{latif_transformers_2023}
Siddique Latif, Aun Zaidi, Heriberto Cuayahuitl, Fahad Shamshad, Moazzam Shoukat, and Junaid Qadir.
\newblock Transformers in {Speech} {Processing}: {A} {Survey}, March 2023.
\newblock arXiv:2303.11607 [cs, eess].

\bibitem{mamatov_speech_2021}
N.~S. Mamatov, N.~A. Niyozmatova, Sh.~Sh. Abdullaev, A.~N. Samijonov, and K.~K. Erejepov.
\newblock Speech {Recognition} {Based} {On} {Transformer} {Neural} {Networks}.
\newblock In {\em 2021 {International} {Conference} on {Information} {Science} and {Communications} {Technologies} ({ICISCT})}, pages 1--5, November 2021.

\bibitem{li_neural_2019}
Naihan Li, Shujie Liu, Yanqing Liu, Sheng Zhao, and Ming Liu.
\newblock Neural speech synthesis with transformer network.
\newblock In {\em Proceedings of the {Thirty}-{Third} {AAAI} {Conference} on {Artificial} {Intelligence} and {Thirty}-{First} {Innovative} {Applications} of {Artificial} {Intelligence} {Conference} and {Ninth} {AAAI} {Symposium} on {Educational} {Advances} in {Artificial} {Intelligence}}, {AAAI}'19/{IAAI}'19/{EAAI}'19, pages 6706--6713, Honolulu, Hawaii, USA, January 2019. AAAI Press.

\bibitem{berroukham_vision_2023}
Abdelhafid Berroukham, Khalid Housni, and Mohammed Lahraichi.
\newblock Vision {Transformers}: {A} {Review} of {Architecture}, {Applications}, and {Future} {Directions}.
\newblock In {\em 2023 7th {IEEE} {Congress} on {Information} {Science} and {Technology} ({CiSt})}, pages 205--210, December 2023.
\newblock ISSN: 2327-1884.

\bibitem{xu_language_2024}
Zelai Xu, Chao Yu, Fei Fang, Yu~Wang, and Yi~Wu.
\newblock Language {Agents} with {Reinforcement} {Learning} for {Strategic} {Play} in the {Werewolf} {Game}, February 2024.
\newblock arXiv:2310.18940 [cs] version: 3.

\bibitem{hu_survey_2024}
Sihao Hu, Tiansheng Huang, Fatih Ilhan, Selim Tekin, Gaowen Liu, Ramana Kompella, and Ling Liu.
\newblock A {Survey} on {Large} {Language} {Model}-{Based} {Game} {Agents}, April 2024.
\newblock arXiv:2404.02039 [cs] version: 1.

\bibitem{liu_large_2024}
Yang Liu, Peng Sun, and Hang Li.
\newblock Large {Language} {Models} as {Agents} in {Two}-{Player} {Games}, February 2024.
\newblock arXiv:2402.08078 [cs].

\bibitem{buehler_fieldperceiver_2022}
Markus~J. Buehler.
\newblock {FieldPerceiver}: {Domain} agnostic transformer model to predict multiscale physical fields and nonlinear material properties through neural ologs.
\newblock {\em Materials Today}, 57:9--25, July 2022.

\bibitem{moussad_transformative_2023}
Bernard Moussad, Rahmatullah Roche, and Debswapna Bhattacharya.
\newblock The transformative power of transformers in protein structure prediction.
\newblock {\em Proceedings of the National Academy of Sciences of the United States of America}, 120(32):e2303499120, August 2023.

\bibitem{brugiere_transformer_2024}
Pierre Brugiere and Gabriel Turinici.
\newblock Transformer for {Times} {Series}: an {Application} to the {S}\&{P500}, March 2024.
\newblock arXiv:2403.02523 [cs, q-fin, stat].

\bibitem{lezmi_time_2023}
Edmond Lezmi and Jiali Xu.
\newblock Time {Series} {Forecasting} with {Transformer} {Models} and {Application} to {Asset} {Management}, February 2023.

\bibitem{vaswani_attention_2023}
Ashish Vaswani, Noam Shazeer, Niki Parmar, Jakob Uszkoreit, Llion Jones, Aidan~N. Gomez, Lukasz Kaiser, and Illia Polosukhin.
\newblock Attention {Is} {All} {You} {Need}, August 2023.
\newblock arXiv:1706.03762 [cs].

\bibitem{saeedizade_navigating_2024}
Mohammad~Javad Saeedizade and Eva Blomqvist.
\newblock Navigating {Ontology} {Development} with {Large} {Language} {Models}.
\newblock In Albert Meroño~Peñuela, Anastasia Dimou, Raphaël Troncy, Olaf Hartig, Maribel Acosta, Mehwish Alam, Heiko Paulheim, and Pasquale Lisena, editors, {\em The {Semantic} {Web}}, pages 143--161, Cham, 2024. Springer Nature Switzerland.

\bibitem{buehler_generative_2024}
Markus~J. Buehler.
\newblock Generative {Retrieval}-{Augmented} {Ontologic} {Graph} and {Multiagent} {Strategies} for {Interpretive} {Large} {Language} {Model}-{Based} {Materials} {Design}.
\newblock {\em ACS Engineering Au}, 4(2):241--277, April 2024.
\newblock Publisher: American Chemical Society.

\bibitem{buehler_accelerating_2024}
Markus~J Buehler.
\newblock Accelerating {Scientific} {Discovery} with {Generative} {Knowledge} {Extraction}, {Graph}-{Based} {Representation}, and {Multimodal} {Intelligent} {Graph} {Reasoning}.
\newblock {\em Machine Learning: Science and Technology}, 2024.

\bibitem{luu_bioinspiredllm_2023}
Rachel~K. Luu and Markus~J. Buehler.
\newblock {BioinspiredLLM}: {Conversational} {Large} {Language} {Model} for the {Mechanics} of {Biological} and {Bio}‐{Inspired} {Materials}.
\newblock {\em Advanced Science}, 11(10):2306724, December 2023.

\bibitem{dobson_reading_2023}
James~E. Dobson.
\newblock On reading and interpreting black box deep neural networks.
\newblock {\em International Journal of Digital Humanities}, 5(2):431--449, November 2023.

\bibitem{bedau_artificial_2007}
Mark~A. Bedau.
\newblock {ARTIFICIAL} {LIFE}.
\newblock In Mohan Matthen and Christopher Stephens, editors, {\em Philosophy of {Biology}}, Handbook of the {Philosophy} of {Science}, pages 585--603. North-Holland, Amsterdam, January 2007.

\bibitem{zhao_progressive_2013}
Yu~Zhao, Fuji Sakai, Lu~Su, Yijiang Liu, Kongchang Wei, Guosong Chen, and Ming Jiang.
\newblock Progressive {Macromolecular} {Self}-{Assembly}: {From} {Biomimetic} {Chemistry} to {Bio}-{Inspired} {Materials}.
\newblock {\em Advanced Materials}, 25(37):5215--5256, 2013.
\newblock \_eprint: https://onlinelibrary.wiley.com/doi/pdf/10.1002/adma.201302215.

\bibitem{capra_complexity_2005}
Fritjof Capra.
\newblock Complexity and {Life}.
\newblock {\em Theory, Culture \& Society}, 22(5):33--44, October 2005.
\newblock Publisher: SAGE Publications Ltd.

\bibitem{skinner_application_1996}
James~E. Skinner, Stewart~G. Wolf, J.~Yasha Kresh, Igor Izrailtyan, J.~A. Armour, and Ming-he Huang.
\newblock Application of chaos theory to a model biological system: {Evidence} of self-organization in the intrinsic cardiac nervous system.
\newblock {\em Integrative Physiological and Behavioral Science}, 31(2):122--146, April 1996.

\bibitem{kurakin_self-organizing_2011}
Alexei Kurakin.
\newblock The self-organizing fractal theory as a universal discovery method: the phenomenon of life.
\newblock {\em Theoretical Biology and Medical Modelling}, 8(1):4, March 2011.

\bibitem{nisioti_text_2024}
Eleni Nisioti, Claire Glanois, Elias Najarro, Andrew Dai, Elliot Meyerson, Joachim~Winther Pedersen, Laetitia Teodorescu, Conor~F. Hayes, Shyam Sudhakaran, and Sebastian Risi.
\newblock From {Text} to {Life}: {On} the {Reciprocal} {Relationship} between {Artificial} {Life} and {Large} {Language} {Models}, June 2024.
\newblock arXiv:2407.09502 [cs].

\bibitem{wolfram_whats_2024}
Stephen Wolfram.
\newblock What’s {Really} {Going} {On} in {Machine} {Learning}? {Some} {Minimal} {Models}, August 2024.

\bibitem{wulff_learning_1992}
N.~Wulff and J~A Hertz.
\newblock Learning {Cellular} {Automaton} {Dynamics} with {Neural} {Networks}.
\newblock In {\em Advances in {Neural} {Information} {Processing} {Systems}}, volume~5. Morgan-Kaufmann, 1992.

\bibitem{mordvintsev_growing_2020}
Alexander Mordvintsev, Ettore Randazzo, Eyvind Niklasson, and Michael Levin.
\newblock Growing {Neural} {Cellular} {Automata}.
\newblock {\em Distill}, 5(2):e23, February 2020.

\bibitem{israeli_coarse-graining_2006}
Navot Israeli and Nigel Goldenfeld.
\newblock Coarse-graining of cellular automata, emergence, and the predictability of complex systems.
\newblock {\em Physical Review E}, 73(2):026203, February 2006.
\newblock arXiv:nlin/0508033.

\bibitem{wolfram_concept_2021}
Stephen Wolfram.
\newblock The {Concept} of the {Ruliad}—{Stephen} {Wolfram} {Writings}, November 2021.

\bibitem{komarov_concept_2003}
A.~S Komarov, M.~M Palenova, and O.~V Smirnova.
\newblock The concept of discrete description of plant ontogenesis and cellular automata models of plant populations.
\newblock {\em Ecological Modelling}, 170(2):427--439, December 2003.

\bibitem{balzter_cellular_1998}
Heiko Balzter, Paul~W. Braun, and Wolfgang Köhler.
\newblock Cellular automata models for vegetation dynamics.
\newblock {\em Ecological Modelling}, 107(2):113--125, April 1998.

\bibitem{halley_competition_1994}
John~M. Halley, Hugh~N. Comins, J.~H. Lawton, and M.~P. Hassell.
\newblock Competition, {Succession} and {Pattern} in {Fungal} {Communities}: {Towards} a {Cellular} {Automaton} {Model}.
\newblock {\em Oikos}, 70(3):435--442, 1994.
\newblock Publisher: [Nordic Society Oikos, Wiley].

\bibitem{ferreira_modelling_2013}
Iuri Emmanuel de~Paula Ferreira, Rafael de~Andrade Moral, Cláudia~Pio Ferreira, and Wesley Augusto~Conde Godoy.
\newblock Modelling fungus dispersal scenarios using cellular automata.
\newblock {\em Ecological Informatics}, 14:53--58, March 2013.

\bibitem{laszlo_cellular_1993}
Joseph~A. Laszlo and Robert~W. Silman.
\newblock Cellular automata simulations of fungal growth on solid substrates.
\newblock {\em Biotechnology Advances}, 11(3):621--633, January 1993.

\bibitem{dobrescu_emergence_2011}
R~Dobrescu and VL~Purcarea.
\newblock Emergence, self–organization and morphogenesis in biological structures.
\newblock {\em Journal of Medicine and Life}, 4(1):82--90, February 2011.

\bibitem{jakab_engineering_2004}
Karoly Jakab, Adrian Neagu, Vladimir Mironov, Roger~R. Markwald, and Gabor Forgacs.
\newblock Engineering biological structures of prescribed shape using self-assembling multicellular systems.
\newblock {\em Proceedings of the National Academy of Sciences}, 101(9):2864--2869, March 2004.
\newblock Publisher: Proceedings of the National Academy of Sciences.

\bibitem{xiao_dna_2018}
Mingshu Xiao, Wei Lai, Xiwei Wang, Xiangmeng Qu, Li~Li, and Hao Pei.
\newblock {DNA} mediated self-assembly of multicellular microtissues.
\newblock {\em Microphysiological Systems}, 2(1), January 2018.
\newblock Number: 1 Publisher: AME Publishing Company.

\bibitem{newman_before_2006}
Stuart~A. Newman, Gabor Forgacs, and Gerd~B. Muller.
\newblock Before programs: the physical origination of multicellular forms.
\newblock {\em The International Journal of Developmental Biology}, 50(2-3):289--299, 2006.

\bibitem{rendell_game_2016}
Paul Rendell.
\newblock Game of {Life} {Universal} {Turing} {Machine}.
\newblock In Paul Rendell, editor, {\em Turing {Machine} {Universality} of the {Game} of {Life}}, pages 71--89. Springer International Publishing, Cham, 2016.

\bibitem{delvenne_computational_2005}
Jean-Charles Delvenne, Petr Kůrka, and Vincent~D. Blondel.
\newblock Computational {Universality} in {Symbolic} {Dynamical} {Systems}.
\newblock In Maurice Margenstern, editor, {\em Machines, {Computations}, and {Universality}}, pages 104--115, Berlin, Heidelberg, 2005. Springer.

\bibitem{reprintsev_turing_2018}
Alex Reprintsev.
\newblock Turing {Completeness}.
\newblock In Alex Reprintsev, editor, {\em Oracle {SQL} {Revealed}: {Executing} {Business} {Logic} in the {Database} {Engine}}, pages 235--242. Apress, Berkeley, CA, 2018.

\bibitem{perez_turing_2019}
Jorge Pérez, Javier Marinković, and Pablo Barceló.
\newblock On the {Turing} {Completeness} of {Modern} {Neural} {Network} {Architectures}, January 2019.
\newblock arXiv:1901.03429 [cs, stat].

\bibitem{perez_attention_2021}
Jorge Pérez, Pablo Barceló, and Javier Marinkovic.
\newblock Attention is {Turing}-{Complete}.
\newblock {\em Journal of Machine Learning Research}, 22(75):1--35, 2021.

\bibitem{schuurmans_autoregressive_2024}
Dale Schuurmans, Hanjun Dai, and Francesco Zanini.
\newblock Autoregressive {Large} {Language} {Models} are {Computationally} {Universal}, October 2024.
\newblock arXiv:2410.03170 [cs].

\bibitem{bhattacharya_strategies_2024}
Ranjeeta Bhattacharya.
\newblock Strategies to mitigate hallucinations in large language models.
\newblock {\em Applied Marketing Analytics}, 10(1):62--67, June 2024.

\bibitem{hamid_beyond_2024}
Oussama~H. Hamid.
\newblock Beyond {Probabilities}: {Unveiling} the {Delicate} {Dance} of {Large} {Language} {Models} ({LLMs}) and {AI}-{Hallucination}.
\newblock In {\em 2024 {IEEE} {Conference} on {Cognitive} and {Computational} {Aspects} of {Situation} {Management} ({CogSIMA})}, pages 85--90, May 2024.

\bibitem{ha_world_2018}
David Ha and Jürgen Schmidhuber.
\newblock World {Models}.
\newblock {\em CoRR}, abs/1803.10122, 2018.
\newblock arXiv: 1803.10122.

\bibitem{wang_lucidrainsx-transformers_2024}
Phil Wang.
\newblock lucidrains/x-transformers, July 2024.
\newblock original-date: 2020-10-24T22:13:25Z.

\bibitem{shannon_mathematical_1948}
C.~E. Shannon.
\newblock A {Mathematical} {Theory} of {Communication}.
\newblock {\em Bell System Technical Journal}, 27(3):379--423, 1948.
\newblock \_eprint: https://onlinelibrary.wiley.com/doi/pdf/10.1002/j.1538-7305.1948.tb01338.x.

\bibitem{mackay_information_2003}
David J.~C. MacKay.
\newblock {\em Information {Theory}, {Inference} and {Learning} {Algorithms}}.
\newblock Cambridge University Press, Cambridge, illustrated edition edition, October 2003.

\bibitem{liu_towards_2023}
Hao Liu, Xinyang Geng, Lisa Lee, Igor Mordatch, Sergey Levine, Sharan Narang, and Pieter Abbeel.
\newblock Towards {Better} {Few}-{Shot} and {Finetuning} {Performance} with {Forgetful} {Causal} {Language} {Models}, January 2023.
\newblock arXiv:2210.13432 [cs].

\end{thebibliography}


\end{document}